\documentclass[10pt,twocolumn,letterpaper]{article}

\usepackage{iccv}
\usepackage{times}
\usepackage{epsfig}
\usepackage{graphicx,subcaption}
\usepackage{amsmath}
\usepackage{amssymb}
\usepackage{booktabs}                      % Professional tables
\usepackage[square,numbers]{natbib}

\usepackage{multirow}
\usepackage{bm}
\usepackage[pagebackref=true,breaklinks=true,letterpaper=true,colorlinks,bookmarks=false]{hyperref}

\iccvfinalcopy % *** Uncomment this line for the final submission

\def\iccvPaperID{2547} % *** Enter the ICCV Paper ID here

\ificcvfinal\fi
\begin{document}

%%%%%%%%% TITLE
\title{Dual Adversarial Inference for Text-to-Image Synthesis}

\author{Qicheng Lao\textsuperscript{\rm 1 2} \qquad Mohammad Havaei\textsuperscript{\rm 1} \qquad Ahmad Pesaranghader\textsuperscript{\rm 1 3} \qquad Francis Dutil\textsuperscript{\rm 1} \qquad \\ 
Lisa Di Jorio\textsuperscript{\rm 1} \qquad Thomas Fevens\textsuperscript{\rm 2} \\
\hspace{-1.25cm}\textsuperscript{\rm 1}Imagia Inc. \qquad \textsuperscript{\rm 2}Concordia University \qquad \textsuperscript{\rm 3}Dalhousie University\\
% {\tt\small \{qi\_lao, fevens\}@encs.concodia.ca, \{mohammad, ahmad.pesaranghader, francis.dutil, lisa\}@imagia.com}
{\tt\small \{qi\_lao, fevens\}@encs.concodia.ca, \{mohammad, ahmad.pgh, francis.dutil, lisa\}@imagia.com}
}

\maketitle
%\thispagestyle{empty}
\ificcvfinal\thispagestyle{empty}\fi

%%%%%%%%% ABSTRACT
\begin{abstract}
Synthesizing images from a given text description involves engaging two types of information: the content, which includes information explicitly described in the text (\eg, color, composition, etc.), and the style, which is usually not well described in the text (\eg, location, quantity, size, etc.). However, in previous works, it is typically treated as a process of generating images only from the content, \ie, without considering learning meaningful style representations. %, without considering the flexibility of using style information.
In this paper, we aim to learn two variables that are disentangled in the latent space, representing content and style respectively. We achieve this by augmenting current text-to-image synthesis frameworks with a dual adversarial inference mechanism. 
Through extensive experiments, we show that our model learns, in an unsupervised manner, style representations corresponding to certain meaningful information present in the image that are not well described in the text. The new framework also improves the quality of synthesized images when evaluated on Oxford-102, CUB and COCO datasets.%, compared to the baseline method that is without inference.

\end{abstract}

%%%%%%%%% BODY TEXT
\section{Introduction}
The problem of text-to-image synthesis is to generate diverse yet plausible images given a text description of the image and a general data distribution of images and matching descriptions.
%Text to image synthesis is the process which we are given a description of an image and the task is to generate diverse but plausible images under the text description and the distribution of the data. 
In recent years, generative adversarial networks (GANs)~\cite{goodfellow2014generative} have asserted themselves as perhaps the most effective architecture for image generation, along with their variant Conditional GANs~\cite{mirza2014conditional}, wherein the generator is conditioned on a vector encompassing some desired property of the generated image.

\begin{figure}[t]
	\centering
	\includegraphics[width=0.47\textwidth]{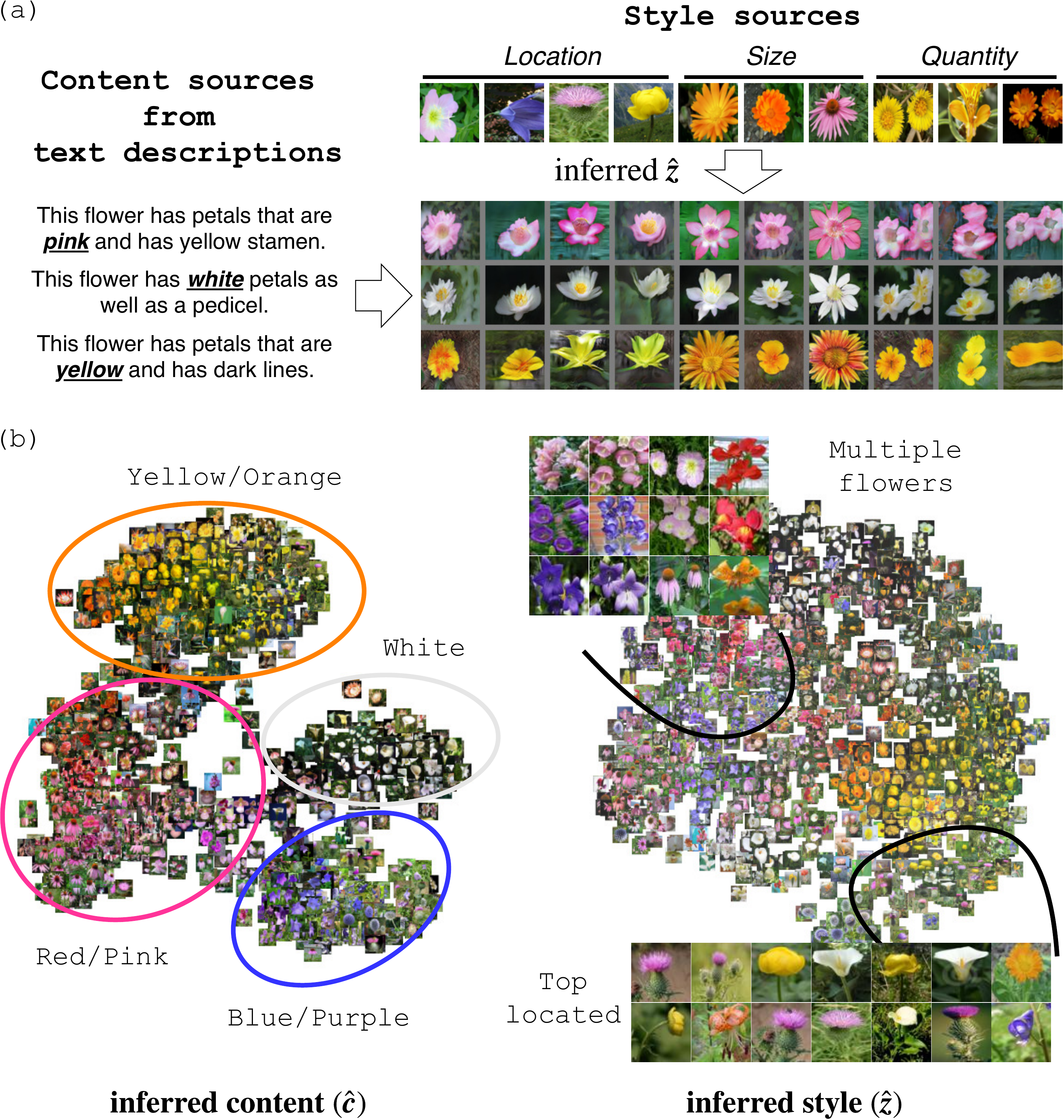}
	\caption{(a) Controlling the style (in columns) of generated images given a text description as the content (in rows). Columns 1-4 show locations (\eg, left, right and top) of the content in the image; Columns 5-7 and columns 8-10 represent size and quantity of the content respectively. (b) The learned content and style features through our dual adversarial inference, visualized by t-SNE. The inferred content is clustered solely on color (one dominant factor that is described in the text), while the inferred style shows a more diffused cluster pattern, with local clusters such as multiple flowers and top-located flowers.}
	\label{fig_first}
% 	\vspace{-3mm}%Put here to reduce too much white space after your table 
\end{figure}

A common approach for text-to-image synthesis is to use a pre-trained text encoder to produce a text embedding from the description. This vector is used as the conditioning factor in a conditional GAN-based model. %To introduce variability in the generated images,
The very first GAN model for the text-to-image synthesis task~\cite{reed2016generative} uses a noise vector sampled from a normal distribution to capture image style features left out of the text representation, 
%to control the randomness of the model to
enabling the model to generate a variety of images given a certain textual description. StackGan~\cite{zhang2017stackgan} introduces conditioning augmentation as a way to augment the text embeddings, where a text embedding can be sampled from a learned distribution representing the text embedding space. As a result, current state-of-the-art methods for text-to-image synthesis generally have two sources of randomness: one for the text embedding variability, and the other (noise $\bm{z}$ given a normal distribution) capturing image variability.

Having two sources of randomness is, however, only meaningful if they represent different factors of variation. 
Problematically, our empirical investigation of some previously published methods reveals that those two sources can overlap: due to the randomness in the text embedding, the noise vector $\bm{z}$ then does not meaningfully contribute to the variability nor the quality of generated images, and can be discarded. This is illustrated in Figure~\ref{fig_problem} and Figure~\ref{fig_is_fid} in the supplementary material.

In this paper we aim to learn a latent space that represents meaningful information in the context of text-to-image synthesis. To do this, we incorporate an inference mechanism that encourages the latent space to learn the distribution of the data. %This allows our latent space to better capture the variability of our data and mitigate the mode collapse problem seen in GAN models~\cite{bang2018improved}. 
To capture different factors of variation, we construct the latent space through two independent random variables, representing content (`$\bm{c}$') and style (`$\bm{z}$'). Similar to previous work~\cite{reed2016generative}, `$\bm{c}$' encodes image content which is the information in the text description. This mostly includes color, composition, etc. On the other hand, `$\bm{z}$' encodes style which we define as all other information in the image data that is not well described in the text. This would typically include location, size, pose, and quantity of the content in the image, background, etc. This new framework allows us to better represent information found in both text and image modalities, achieving better results on Oxford-102~\cite{nilsback2008automated}, CUB~\cite{welinder2010caltech} and COCO~\cite{lin2014microsoft} datasets at 64$\times$64 resolution. 

%The main goal of this paper is to show how unsupervised representation learning of content and style can improve text-to-image synthesis and to that end %Note that in this work, 
%we only focus on the generation of low-resolution images (\ie, 64$\times$64).  
The main goal of this paper is to learn disentangled representations of style and content through an inference mechanism for text-to-image synthesis. This allows us to use not only the content information described in the text descriptions but also the desired styles when generating images. To that end, we only focus on the generation of low-resolution images (\ie, 64$\times$64). %unsupervised representation of 
%which we think is the most essential part of text-to-image synthesis. 
In the literature, high-resolution images are generally produced by iterative refinement of lower-resolution images and thus we consider it a different task, more closely related to generating super-resolution images. % It is not the focus of the present work. 
%However, since the first work that has been done at 64$\times$64 resolution \cite{reed2016generative}, most recent works have been migrated into different scales of resolution: 128$\times$128 \cite{reed2016learning2, hong2018inferring} and 256$\times$256 \cite{zhang2017stackgan, dash2017tac, zhang2018photographic, xu2017attngan}, which are actually composite of two different tasks: the text-to-image synthesis task itself followed by a super resolution task conditioned on the text description again. We strongly encourage the community to take this into consideration for future work and separate low-resolution image generation from high-resolution image generation in order for different methods to be comparable.

To the best of our knowledge, this is the first time an attempt has been made to explicitly separate the learning of style and content for text-to-image synthesis. We believe that capturing these subtleties is important to learn richer representations of the data.  %For instance, the same content can appear in many different styles and ideally we would like to allow for controlling the style while keeping the content constant (see Figure~\ref{fig_first} (a)). As shown in Figure~\ref{fig_first} (b), we learn two disentangled representations of content and style. 
As shown in Figure~\ref{fig_first}, by learning disentangled representations of content and style, we can generate images that respect the content information from a text source while controlling style by inferring the style information from a style source. %Note that in this paper, the disentanglement is defined as the separation of content and style, \ie, the disentanglement of style with respect to content.
It is worth noting that although we hope to learn the style from the image modality, the style information could possibly be connected to (or leaked into) some text instances. Despite this, the integration of the style in the model eventually depends on how well it is represented in both modalities. %how well the model captures that information in both modalities that are generalizable. 
For example, if certain types of style information are commonly present in the text, then according to our definition, those types of information are considered as content. If only a few text instances describe that information however, then it would not be fully representative of a shared commonality among texts and therefore would not be captured as content, and whether it can be captured as style depends on how well it is represented in the image modality. On the other hand, we would also like to explore modalities other than \textit{text} as the content in our future work using the proposed method, which may bring us closer to image-to-image translation~\cite{lee2018diverse} if we choose both modalities to be \textit{image}.

The contributions of this paper are twofold: $(i)$~we are the first to learn two variables that are disentangled for content and style in the context of text-to-image synthesis using inference; and $(ii)$ by incorporating inference we improve on the state-of-the-art in image quality while maintaining comparable variability and visual-semantic similarity when evaluated on the Oxford-102, CUB and COCO datasets.

\begin{figure*}[htbp]
	\centering
	\includegraphics[width=\textwidth]{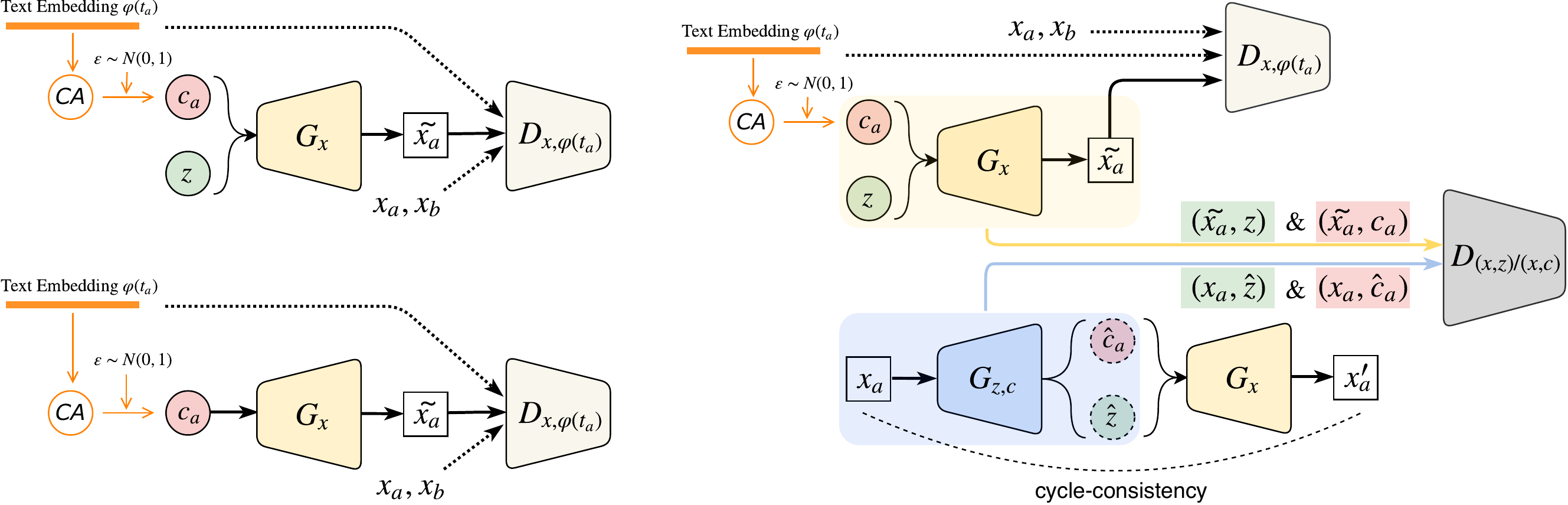}
	\caption{Overview of the current state-of-the-art methods (left top) and our proposed method (right) for text-to-image synthesis at low-resolution scale. By default, the current state-of-the-art methods adopt \emph{conditioning augmentation} (CA), which introduces variable $\bm{c} \sim p(\bm{c}|\varphi_t)$, in addition to variable $\bm{z} \sim \mathcal{N}(0,1)$ as the inputs for the image generator $G_x$.
	%where variable $\bm{c}$ is sampled from a learned distribution through $\varphi_t$ and then concatenated with variable $\bm{z}$ sampled from a normal distribution as the input for the image generator $G_x$. 
	The removal of $\bm{z}$ (left bottom) does not affect the model performance (\emph{viz.} Figure~\ref{fig_is_fid} in supplementary material for quantitative evaluations). In our method (right), we incorporate the inference mechanism, where $G_{z,c}$ encodes both $\bm{z}$ and $\bm{c}$, and the discriminator $D_{(x,z)/(x,c)}$ distinguishes between joint pairs. For the cycle consistency, sampled $\hat{\bm{z}}$ and $\hat{\bm{c}}$ are also used to reconstruct $\bm{x}'$.}
	\label{fig2_architecture}
	\vspace{-0.4mm}%Put here to reduce too much white space after your table 
\end{figure*}

%------------------------------------------------------------------------
\section{Related Work}
%This section describes the connections of our work to previous related work.

\paragraph{Text-to-image synthesis methods}
Text-to-image synthesis has been made possible by Reed \etal \cite{reed2016generative}, where a conditional GAN-based model is used to generate text-matching images from the text description. Zhang \etal~\cite{zhang2017stackgan} use a two-stage GAN to first generate low-resolution images in stage I and then improve the image quality to high-resolution in stage II. By using a hierarchically-nested GAN (HDGAN) which incorporates multiple loss functions at increasing levels of resolution, Zhang \etal~\cite{zhang2018photographic} further improve the state-of-the-art on this task in an end-to-end manner. Several attempts have been made to leverage additional available information, such as object location \cite{reed2016learning2}, class label \cite{dash2017tac, cha2018adversarial}, attention extracted from word features \cite{xu2017attngan, qiao2019mirrorgan} and text regeneration \cite{qiao2019mirrorgan}. Hong \etal \cite{hong2018inferring} propose another approach by providing the image generator with a semantic structure that is sequentially constructed with a box generator followed by a shape generator; however, their approach would not be applicable for single-object image synthesis. 
Compared to all previous work, our method incorporates the inference mechanism into the current framework for text-to-image synthesis, and by doing so, we explicitly force the model to simultaneously learn separate representations of content and style. 

Reed \etal \cite{reed2016generative} have also investigated the separation of content and style information. %The differences are elaborated in the supplementary material. 
However, their learning of style is detached from the text-to-image framework, and the parameters of the image generator are fixed during the style learning phase. Therefore, their concept of content and style separation is not actually leveraged during the training of the image generator. In addition, their work uses a deterministic text embedding, which cannot plausibly cover all content variations, and as a result, one can assume that information belonging to the content could severely contaminate the style. In our work, we learn style from the data itself as opposed to the generated images. This allows us to learn style while updating the generator and effectively incorporate style information from the data into the generator. 

%Reed \etal \cite{reed2016generative} have also investigated the separation of content and style information. However in their framework, the style encoder is learned from the model distribution after the image generator is trained, which may not recover the style information from the data distribution. In our case, however, 

\paragraph{Adversarial inference methods}
Various papers have explored learning representations through adversarial training. Notable mentions are BiGANs~\cite{donahue2017adversarial,dumoulin2017adversarially} where a bidirectional discriminator acts on pairs $\bm{(x,z)}$ of data and generated points. While these models assume that a single random variable $\bm{z}$ encodes data representations, in this work we extend the adversarial inference to two random variables that are disentangled with each other. 
Our model is also closely related to~\cite{li2017alice}, where the authors incorporate an adversarial reconstruction loss into the BiGAN framework. They show that the additional loss term results in better reconstructions and more stable training. 
Although Dumoulin \etal ~\cite{dumoulin2017adversarially} show results for conditional image generation, in their model the conditioning factor is discrete, fully observed and not inferred through the inference model. %Also in ali the $y$ variable is binary and fully factorized and each dimension controls for a particular variable. 
In our model however, `$\bm{c}$' can be a continuous conditioning variable that we infer from the text and image. %we do try to make it factorized via KL on the c distribution. Also the ali loss would affect the text embedding encoder through $G_x$. 

\paragraph{Relation to InfoGAN}
While the matching-aware loss (Section~\ref{sec_methods_prelim}) used in many text-to-image works can also be viewed as maximizing mutual information between the two modalities %multi-modalities
(\ie, \textit{text} and \textit{image}), the way it is approximated is different. InfoGAN~\cite{chen2016infogan} uses the variational mutual information maximization technique, whereas the matching-aware loss uses the concept of matched and mismatched pairs. In addition, InfoGAN concentrates all semantic features on the latent code $\bm{c}$, which contains both content and style, whereas in this work, we only maximize mutual information on the content since we consider \textit{text} as our content.

%------------------------------------------------------------------------
\section{Methods}
\subsection{Preliminaries} \label{sec_methods_prelim}
We start by describing text-to-image synthesis. 
Let ${\varphi_t}$ be the text embedding of a given text description associated with image $\bm{x}$. The goal of text-to-image synthesis is to generate a variety of visually-plausible images that are text-matched. Reed $\etal$ \cite{reed2016generative} first propose a conditional GAN-based framework, where a generator $G_x$ takes as input a noise vector $\bm{z}$ sampled from $p(\bm{z}) = \mathcal{N}(0, 1)$ and ${\varphi_t}$ as the conditioning factor to generate an image $\tilde{\bm{x}} = G_x(\bm{z}, {\varphi_t})$. A {\it matching-aware} discriminator $D_{x, \varphi_t}$ is then trained to not only judge between real and fake images, but also discriminate between matched and mismatched image-text pairs. The minimax objective function for text-to-image (subscript denoted as $t2i$) framework is given as:
\begin{multline*}\label{eq:original_gan_obj} \tag{1}
    \min_G \max_D V_{t2i}(D_{x, \varphi_t}, G_x) = \\
    \mathbb{E}_{(\bm{x}_a, {t}_a) \sim p_\text{data}}[\log D_{x, \varphi_t}(\bm{x}_a, {\varphi_{t_a}})] + \\ 
    \frac{1}{2} \big\{ \mathbb{E}_{(\bm{x}_a, t_b) \sim p_\text{data}}[\log(1 - D_{x, \varphi_t}(\bm{x}_a, \varphi_{t_b}))] + \\ 
    \mathbb{E}_{\bm{z} \sim p(\bm{z}), t_a \sim p_\text{data}}[\log(1 - D_{x, \varphi_t}(G_x(\bm{z}, \varphi_{t_a}), \varphi_{t_a}))] \big\},
\end{multline*}
where $(\bm{x}_a, t_a)$ is a matched pair and $(\bm{x}_a, t_b)$ is a mismatched pair.

To augment the text data, Zhang \etal \cite{zhang2017stackgan} replace the deterministic text embedding $\varphi_t$ in the generator with a latent variable $\bm{c}$, which is sampled from a learned Gaussian distribution $p(\bm{c}|\varphi_t) = \mathcal{N}(\mu(\varphi_t),\,\Sigma(\varphi_t))$, where $\mu$ and $\Sigma$ are functions of $\varphi_t$ parameterized by neural networks. For simplicity in notation, we denote $p(\bm{c}|\varphi_t)$ as $p(\bm{c})$. As a result, the objective function (\ref{eq:original_gan_obj}) is updated to:
\begin{multline*}\label{eq:stackgan_obj} \tag{2}
    \min_G \max_D V_{t2i}(D_{x, \varphi_t}, G_x) = \\
    \mathbb{E}_{(\bm{x}_a, t_a) \sim p_\text{data}}[\log D_{x, \varphi_t}(\bm{x}_a, \varphi_{t_a})] + \\ 
    \frac{1}{2} \big\{ \mathbb{E}_{(\bm{x}_a, t_b) \sim p_\text{data}}[\log(1 - D_{x, \varphi_t}(\bm{x}_a, \varphi_{t_b}))] + \\ 
    \mathbb{E}_{\bm{z} \sim p(\bm{z}), \textcolor{blue}{\bm{c} \sim p(\bm{c})}, t_a \sim p_\text{data}}[\log(1 - D_{x, \varphi_t}(G_x(\bm{z}, \textcolor{blue}{\bm{c}}), \varphi_{t_a}))] \big\}.\\[-3ex]
\end{multline*}

In addition to the matching-aware pair loss that guarantees the semantic consistency, Zhang \etal \cite{zhang2018photographic} propose another type of adversarial loss that focuses on the image fidelity (\ie, image loss), further updating (\ref{eq:stackgan_obj}) to:
\begin{multline*}\label{eq:hdgan_obj} \tag{3}
    \min_G \max_D V_{t2i}(D_x, D_{x, \varphi_t}, G_x) = \\
    \resizebox{\linewidth}{!}{
        $\textcolor{blue}{\mathbb{E}_{\bm{x}_a \sim p_\text{data}}[\log D_x(\bm{x}_a)] + \mathbb{E}_{\bm{z} \sim p(\bm{z}), \bm{c} \sim p(\bm{c}})[\log(1 - D_x(G_x(\bm{z}, \bm{c})))]} + $
    }\\
    \mathbb{E}_{(\bm{x}_a, t_a) \sim p_\text{data}}[\log D_{x, \varphi_t}(\bm{x}_a, \varphi_{t_a})] + \\ 
    \frac{1}{2} \big\{ \mathbb{E}_{(\bm{x}_a, t_b) \sim p_\text{data}}[\log(1 - D_{x, \varphi_t}(\bm{x}_a, \varphi_{t_b}))] + \\
    \mathbb{E}_{\bm{z} \sim p(\bm{z}), \bm{c} \sim p(\bm{c}), t_a \sim p_\text{data}}[\log(1 - D_{x, \varphi_t}(G_x(\bm{z}, \bm{c}), \varphi_{t_a}))] \big\}, \\[-3ex]
\end{multline*}
where $D_x$ is a discriminator distinguishing between images sampled from $p_\text{data}$ and those sampled from the distribution parameterized by the generator (\ie, $p_\text{model}$). 

Consider two general probability distributions $q(\bm{x})$ and $p(\bm{z})$ over two domains $\bm{x} \in \mathcal{X}$ and $\bm{z} \in \mathcal{Z}$, where $q(\bm{x})$ represents the empirical data distribution and $p(\bm{z})$ is usually specified as a simple random distribution, $\eg$, a standard normal $\mathcal{N}(0,1)$. Adversarial inference \cite{donahue2017adversarial,dumoulin2017adversarially} aims to match the two joint distributions $q(\bm{x},\bm{z})=q(\bm{z}|\bm{x})q(\bm{x})$ and $p(\bm{x},\bm{z})=p(\bm{x}|\bm{z})p(\bm{z})$, which in turn implies that $q(\bm{z}|\bm{x})$ matches $p(\bm{z}|\bm{x})$. To achieve this, an encoder $G_z(\bm{x}): \hat{\bm{z}}=G_z(\bm{x}), \bm{x} \sim q(\bm{x})$ is introduced in the generation phase, in addition to the standard generator $G_x(\bm{z}): \tilde{\bm{x}}=G_x(\bm{z}), \bm{z} \sim p(\bm{z})$. The discriminator $D$ is trained to distinguish joint pairs between $(\bm{x}, \hat{\bm{z}})$ and $(\tilde{\bm{x}}, \bm{z})$. The minimax objective of adversarial inference can be written as:
\begin{multline*}\label{eq:ali_obj} \tag{4}
    \min_G \max_D V(D, G_x, G_z) = \\
    \mathbb{E}_{\bm{x} \sim q(\bm{x}), \hat{\bm{z}} \sim q(\bm{z}|\bm{x})}[\log D(\bm{x}, \hat{\bm{z}})] + \\
    \mathbb{E}_{\tilde{\bm{x}} \sim p(\bm{x}|\bm{z}), \bm{z} \sim p(\bm{z})}[\log(1 - D(\tilde{\bm{x}}, \bm{z}))]. \hspace{7.4mm}
\end{multline*}
\begin{figure*}[htbp]
	\centering
	\includegraphics[width=\textwidth]{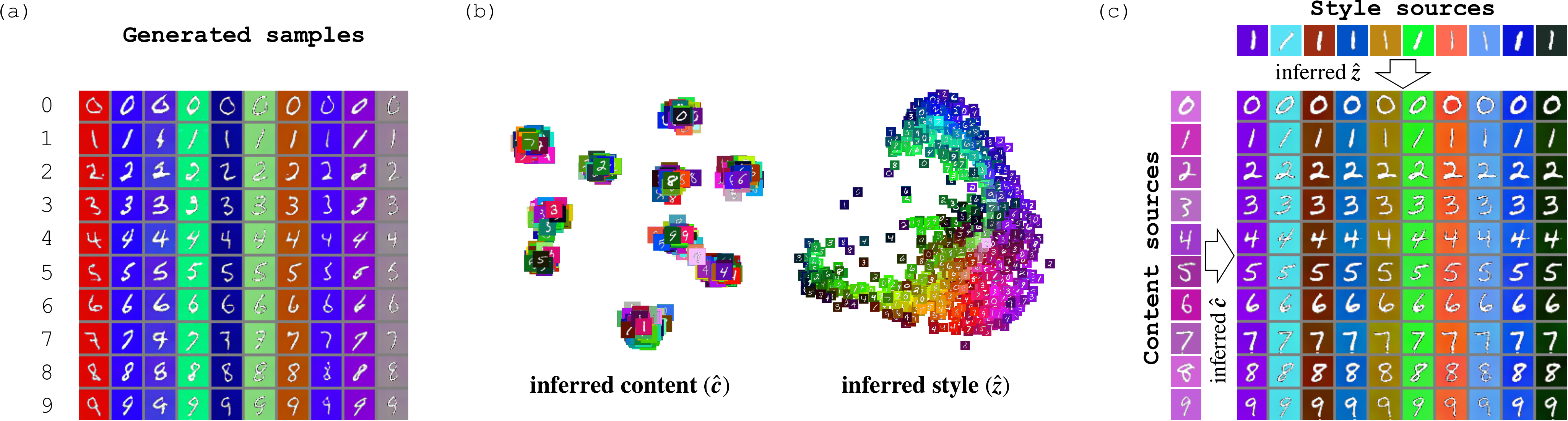}
	\caption{Disentangling content and style on MNIST-CB dataset. (a) Generated samples given digit identities as the content $\bm{c}$. Each column uses the same style $\bm{z}$ sampled from $\mathcal{N}(0,1)$. (b) The t-SNE visualizations of inferred content $\hat{\bm{c}}$ and inferred style $\hat{\bm{z}}$. (c) Reconstructed samples using inferred content $\hat{\bm{c}}$ (in rows) and inferred style $\hat{\bm{z}}$ (in columns) from image sources.}
	\label{fig_mnist}
% 	\vspace{-3mm}%Put here to reduce too much white space after your table 
\end{figure*}
\subsection{Dual adversarial inference}
As described in Section \ref{sec_methods_prelim}, the current state-of-the-art methods for text-to-image synthesis can be viewed as variants of conditional GANs, where the conditioning is initially on $\varphi_t$ itself \cite{reed2016generative} and later on updated to the latent variable $\bm{c}$ sampled from a distribution learned through $\varphi_t$~\cite{zhang2017stackgan, zhang2018photographic, xu2017attngan, qiao2019mirrorgan}. The generator then has two latent variables $\bm{z}$ and $\bm{c}$: $\bm{z} \sim p(\bm{z})$, $\bm{c} \sim p(\bm{c})$ (left, Figure \ref{fig2_architecture}). The priors can be Gaussian or non-Gaussian distributions such as the Bernoulli distribution \footnote{In this paper, we experiment with both Gaussian and Bernoulli distributions for $p(\bm{c})$ (More details in Section \ref{experiments}).}.
%$p(\bm{z}) = \mathcal{N}(0,1), p(\bm{c}) = \mathcal{N}(\mu(\varphi_t),\,\Sigma(\varphi_t))$. %To understand what kind of features $z$ and $c$ are representing and more importantly, to enforce the separation of style features ($z$) and content features ($c$), we incorporate dual adversarial inference learning into the current framework for text-to-image synthesis.
To learn disentangled representations for style ($\bm{z}$) and content ($\bm{c}$) and to enforce the separation between these two variables, we incorporate dual adversarial inference into the current framework for text-to-image synthesis (right, Figure \ref{fig2_architecture}). In this dual inference process, we are interested in matching the conditional $q(\bm{z},\bm{c}|\bm{x})$ to the posterior $p(\bm{z},\bm{c}|\bm{x})$, which under the independence assumption can be factorized as follows:
\begin{align*}
q(\bm{z},\bm{c} \mid \bm{x}) &= q(\bm{z}\mid\bm{x})q(\bm{c}\mid\bm{x}), \\
p(\bm{z},\bm{c} \mid \bm{x}) &= p(\bm{z}\mid\bm{x})p(\bm{c}\mid\bm{x}).
\end{align*}
This formulation allows us to match $q(\bm{z}|\bm{x})$ with $p(\bm{z}|\bm{x})$ and $q(\bm{c}|\bm{x})$ with $p(\bm{c}|\bm{x})$, respectively. Similar to previous work \cite{dumoulin2017adversarially, donahue2017adversarial}, we achieve this by matching the two pairs of joint distributions: 
\begin{align*}
q(\bm{z},\bm{x}) &= p(\bm{z},\bm{x}), \\
q(\bm{c},\bm{x}) &= p(\bm{c},\bm{x}).
\end{align*}
The encoder for our dual adversarial inference then encodes both $\bm{z}$ and $\bm{c}$: $\hat{\bm{z}}, \hat{\bm{c}} = G_{z,c}(\bm{x}), \bm{x} \sim q(\bm{x})$, while the generator decodes $\bm{z}$ and $\bm{c}$ sampled from their corresponding prior distributions into an image: $\tilde{\bm{x}} = G_x(\bm{z},\bm{c}), \bm{z} \sim p(\bm{z}), \bm{c} \sim p(\bm{c})$. To compete with $G_x$ and $G_{z,c}$, the discrimination phase also has two components: the discriminator $D_{x,z}$ is trained to discriminate $(\bm{x},\bm{z})$ pairs sampled from either $q(\bm{x},\bm{z})$ or $p(\bm{x},\bm{z})$, and the discriminator $D_{x,c}$ for the discrimination of $(\bm{x},\bm{c})$ pairs sampled from either $q(\bm{x},\bm{c})$ or $p(\bm{x},\bm{c})$. Given the above setting, the original adversarial inference objective (\ref{eq:ali_obj}) is updated as:
\begin{multline*}\label{eq:dual_obj} \tag{5}
    \min_G \max_D V_{dual}(D_{x,z},D_{x,c}, G_x, G_{z,c}) = \\
    % \mathbb{E}_{x \sim q(x), \hat{z}, \hat{c} \sim q(z,c|x)}[\log(D(x, \hat{z})) + \log(D(x, \hat{c}))] + \\
    % \mathbb{E}_{\tilde{x} \sim p(x|z,c), z \sim p(z), c \sim p(c)}[\log(1 - D(\tilde{x}, z)) + \log(1 - D(\tilde{x}, c))]. \\[-3ex]
    \resizebox{0.8\linewidth}{!}{
        $\mathbb{E}_{\bm{x} \sim q(\bm{x}), \hat{\bm{z}}, \hat{\bm{c}} \sim q(\bm{z},\bm{c}|\bm{x})}[\log D_{x,z}(\bm{x}, \hat{\bm{z}}) + \log D_{x,c}(\bm{x}, \hat{\bm{c}})] +$ 
    }\\
    \resizebox{\linewidth}{!}{
        $\mathbb{E}_{\tilde{\bm{x}} \sim p(\bm{x}|\bm{z},\bm{c}), \bm{z} \sim p(\bm{z}), \bm{c} \sim p(\bm{c})}[\log(1 - D_{x,z}(\tilde{\bm{x}}, \bm{z})) + \log(1 - D_{x,c}(\tilde{\bm{x}}, \bm{c}))]$.
    }\\[-3ex]
\end{multline*}

\subsection{Cycle consistency} \label{sec_method_cycle}
In unsupervised learning, {\it cycle-consistency} refers to the ability of the model to reconstruct the original image $\bm{x}$ from its inferred latent variable $\bm{z}$.
It has been reported that bidirectional adversarial inference models often have difficulties in reproducing faithful reconstructions as they do not explicitly include any reconstruction loss in the objective function \cite{dumoulin2017adversarially, donahue2017adversarial, li2017alice}. The cycle-consistency criterion, as having been demonstrated in many previous works such as CycleGAN \cite{CycleGAN2017}, DualGAN \cite{yi2017dualgan}, DiscoGAN \cite{kim2017learning} and augmented CycleGAN \cite{almahairi2018augmented}, enforces a strong connection between domains (here $\bm{x}$ and $\bm{z}$) by constraining the models (\eg, encoder and decoder) to be consistent with one another.
%Li \etal \cite{li2017alice} have also mathematically proven that cycle-consistency is an upper bound of the conditional entropy, which measures the uncertainty of $x$ from one domain given $z$ to another domain, and vice versa. 
Li \etal ~\cite{li2017alice} show that the integration of %conditional entropy by optimizing 
the cycle-consistency objective stabilizes the learning of adversarial inference, thus yielding better reconstruction results. 

With the above in mind, we integrate cycle-consistency in our dual adversarial inference framework in a similar fashion to~\cite{li2017alice} . More concretely, we use another discriminator $D_{x, x'}$ to distinguish between $\bm{x}$ and its reconstruction $\bm{x}'= G_x(\hat{\bm{z}}, \hat{\bm{c}})$, where $\hat{\bm{z}}, \hat{\bm{c}} = G_{z,c}(\bm{x})$, by optimizing:
\begin{multline*}\label{eq:cycle_obj} \tag{6}
    \min_G \max_D V_{cycle}(D_{x,x'}, G_x, G_{z,c}) = \\
    \mathbb{E}_{\bm{x} \sim q(\bm{x})}[\log D_{x,x'}(\bm{x}, \bm{x})] + \\
    \mathbb{E}_{\bm{x} \sim q(\bm{x}), (\hat{\bm{z}},\hat{\bm{c}}) \sim q(\bm{z},\bm{c}|\bm{x})}[\log(1 - D_{x,x'}(\bm{x}, G_x(\hat{\bm{z}}, \hat{\bm{c}})))]. \\[-3ex]
\end{multline*}
We later show in an ablation study (Section \ref{ablation_study}) that using $l_2$ loss for cycle-consistency leads to blurriness in the generated images, which agrees with previous studies \cite{larsen2016autoencoding, yi2017dualgan}.

\subsection{Full objective}
Taking (\ref{eq:hdgan_obj}), (\ref{eq:dual_obj}), (\ref{eq:cycle_obj}) into account, our full objective is:
\begin{multline*}\label{eq:full_obj} \tag{7}
    % \resizebox{0.9\linewidth}{!}{
    %     $\min \limits_{G} \max \limits_{D} V(D_{x, \varphi_t}, D_{x,z}, D_{x,c}, D_{x,x'}, G_x, G_{z,c})$
    % }\\
    \hspace{8.5mm} \min \limits_{G} \max \limits_{D} V_{full}(D, G) \\
    = V_{t2i}(D_x, D_{x, \varphi_t}, G_x) \hspace{6mm} \\
    \hspace{5mm} + V_{dual}(D_{x,z},D_{x,c}, G_x, G_{z,c}) \\
    + V_{cycle}(D_{x,x'}, G_x, G_{z,c}), \hspace{15mm}
\end{multline*}
where $G$ and $D$ are the sets of all generators and discriminators in our method: $G=\{ G_x, G_{z,c} \}$ and $D=\{D_x, D_{x, \varphi_t}, D_{x,z}, D_{x,c}, D_{x,x'} \}$.

Note that in addition to the latent variable $\bm{c}$, the encoded $\hat{\bm{z}}$ and $\hat{\bm{c}}$ in our method are also sampled from the inferred posterior distributions through the reparameterization trick \cite{kingma2014auto}, \ie, $\hat{\bm{z}} \sim q(\bm{z}|\bm{x})$ and $\hat{\bm{c}} \sim q(\bm{c}|\bm{x})$. In order to encourage smooth sampling over the latent space, we regularize the posterior distributions $q(\bm{z}|\bm{x})$ and $q(\bm{c}|\bm{x})$ to match their respective priors by minimizing the KL divergence. %, \ie, $\lambda (D_{KL}(q(\bm{z}|\bm{x}) \, || \, \mathcal{N}(0, 1)) + D_{KL}(q(\bm{c}|\bm{x}) \, || \, \mathcal{N}(0, 1)) )$. %where $p(\bm{z})$ and $p{\bm{c}}$ are standard normal distributions.
We apply a similar regularization term to $p(\bm{c})$, \eg, $\lambda D_{KL}(p(\bm{c}) \, || \, \mathcal{N}(0, 1))$ for a normal distribution prior, as done in previous text-to-image synthesis works \cite{zhang2017stackgan, zhang2018photographic}.
% \begin{multline*}\label{eq:full_obj_gx} \tag{8}
%     \resizebox{0.96\linewidth}{!}{
%         $\mathcal{L}_{G_x} = \mathbb{E}_{z \sim p_{z}, c \sim p_c, t_a \sim p_\text{data}}[\log(1 - D_{x, \varphi_t}(G_x(z, c), \varphi_{t_a}))]$
%     }\\
%     \resizebox{\linewidth}{!}{
%         $+ \mathbb{E}_{z \sim p_{z}, c \sim p_{c}}[\log(1 - D_x(G_x(z, c)))] + \lambda D_{KL}(p(c) || \mathcal{N}(0,1))$,
%     }\\[-3ex]
% \end{multline*}
Our preliminary experiments \footnote{We also experimented with minimizing the cosine similarity between $\hat{\bm{z}}$ and $\hat{\bm{c}}$, but did not observe improved performance in terms of the inception score and FID.} showed that without the above regularization, the training became unstable and the gradients typically explode after certain number of epochs. 

\begin{table*}[htbp]
    % NOTE by NC: In tables, captions should always be on top of the table. For figures, captions should be below the figure.
    % In CVPR guidelines, the captions of table are recommended at the bottom.
% \renewcommand{\arraystretch}{1.25}
		\begin{tabular}{l@{\hspace{0.9cm}}ccc@{\hspace{0.9cm}}ccc}
			\toprule
			\multirow{2}{*}{Method} & \multicolumn{3}{c@{\hspace{1cm}}}{{Inception Score}} & \multicolumn{3}{c}{{FID}} \\\cmidrule{2-4}\cmidrule{5-7}
			                         & Oxford-102 & CUB &COCO & Oxford-102 & CUB &COCO \\\midrule
			GAN-INT-CLS \cite{reed2016generative} & 2.66 $\pm$ 0.03 & 2.88 $\pm$ 0.04 & 7.88 $\pm$ 0.07 & 79.55 & 68.79 & 60.62\\
			GAWWN \cite{reed2016learning2} & --- & 3.10 $\pm$ 0.03 & --- & --- & 53.51 & --- \\
			StackGAN \cite{zhang2017stackgan, zhang2017stackgan++} & 2.73 $\pm$ 0.03 & 3.02 $\pm$ 0.03 & 8.35 $\pm$ 0.11 & 43.02 & 35.11 & 33.88 \\
			HDGAN \cite{zhang2018photographic} & --- & 3.53 $\pm$ 0.03 & --- & --- & --- & --- \\
			\midrule
			HDGAN mean* & 2.90 $\pm$ 0.03 & \textbf{3.58 $\pm$ 0.03} & 8.64 $\pm$ 0.37 & 40.02 $\pm$ 0.55 & 20.60 $\pm$ 0.96 & 29.13 $\pm$ 3.76 \\
			Ours mean* & \textbf{2.90 $\pm$ 0.03} & \textbf{3.58 $\pm$ 0.05} & \textbf{8.94 $\pm$ 0.20} & \textbf{37.94 $\pm$ 0.39} & \textbf{18.41 $\pm$ 1.07} & \textbf{27.07 $\pm$ 2.55} \\
			\bottomrule
			\multicolumn{7}{l}{\footnotesize * mean calculated on three experiments at five different epochs (600, 580, 560, 540, 520), or three different epochs (200, 190, 180) for COCO dataset }
		  %  \multicolumn{}{}{\footnotesize * mean calculated on three experiments at five different epochs (520, 540, 560, 580 and 600)}
% 			\footnote{for COCO dataset, we use three different epochs (200, 190, 180).
		\end{tabular}
% 	\vspace{-2mm}%Put here to reduce too much white space after your table 
	\caption{Comparison of inception score and FID at 64$\times$64 resolution scale. Higher inception score and lower FID mean better performance.}
	\label{tab1_performance}
% 	\vspace{-3mm}%Put here to reduce too much white space after your table 
\end{table*}

\begin{figure*}[htbp]
	\centering
	\includegraphics[width=\textwidth]{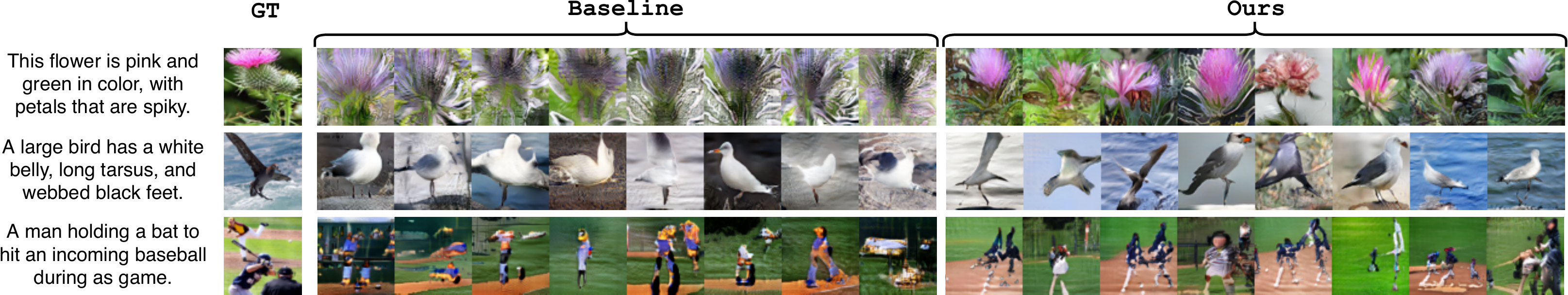}
	\caption{Examples of generated images on Oxford-102 (top), CUB (middle) and COCO (bottom) datasets.}
	\label{fig_generation}
% 	\vspace{-3mm}%Put here to reduce too much white space after your table 
\end{figure*}

%------------------------------------------------------------------------
\section{Experiments} \label{experiments}
\subsection{Proof-of-concept study}
To evaluate the effectiveness of our proposed dual adversarial inference on the disentanglement of content and style, we first validate our proposed method on a toy dataset: MNIST-CB~\cite{gonzalez2018image}, where we formulate %transform
the digit generation problem as %into 
a text-to-image synthesis problem by considering the digit identity as the text content. In this setup, digit font and background color represent styles learned in an unsupervised manner through adversarial inference. We add a cross-entropy regularization term to the content inference objective since our content in this case is discrete (\ie, one-hot vector for digit identity). As shown in Figure~\ref{fig_mnist}~(a), the content and style are disentangled in the generation phase, where the generator has learned to assign the same style to different digit identities when the same $\bm{z}$ is used. More importantly, the t-SNE visualizations (Figure~\ref{fig_mnist}~(b)) from our inferred content and style ($\hat{\bm{c}}$ and $\hat{\bm{z}}$) %show their corresponding cluster patterns in the  (%\ie, clusters of digit identities for $\hat{\bm{c}}$, and clusters of colors and fonts for $\hat{\bm{z}}$
indicate that our dual adversarial inference has successfully separated the information on content (digit identity) and style (font and background color). %This is further validated in Figure~\ref{fig_mnist}(c), where we encode content and style from different sources of test images for the regeneration, and the regenerated images show consistency in respecting both content and style of their sources.
%This is further validated in Figure~\ref{fig_mnist}(c) were we show how the model can fuse style and content inferred from two separate set of images to generate images which use the content from one set and the style from the other.    
This is further validated in Figure~\ref{fig_mnist}~(c) where we show our model's ability to infer style and content from different image sources and fuse them to generate hybrid images, using content from one source and style from the other.
%This is further validated in Figure~\ref{fig_mnist}(c) were we show our model's ability to fuse style and content information inferred from different image sources to generate hybrid images using content from one source and style from the other.

\subsection{Text-to-image setup}
Once validated on the toy example, we move to the original text-to-image synthesis task.
We evaluate our method based on model architectures similar to HDGAN~\cite{zhang2018photographic}, one of the current state-of-the-art methods for text-to-image synthesis, making HDGAN our baseline method. The architecture designs are the same as described in~\cite{zhang2018photographic}, keeping in mind that we only consider the 64$\times$64 resolution. Three quantitative metrics are used to evaluate our method: Inception score \cite{salimans2016improved}, Fr\'echet inception distance (FID)~\cite{heusel2017gans} and Visual-semantic similarity~\cite{zhang2018photographic}. It has been noticed in our experiments and also reported by others \cite{lucic2018gans} that, due to the variations in the training of GAN models, it is unfair to draw a conclusion based on one single experiment that achieves the best result; therefore, in our experiments, we perform three independent experiments for each method, with averages reported as final results. More implementation, dataset and evaluation details can be found in the supplementary material.

\begin{figure*}[htbp]
	\centering
	\includegraphics[width=\textwidth]{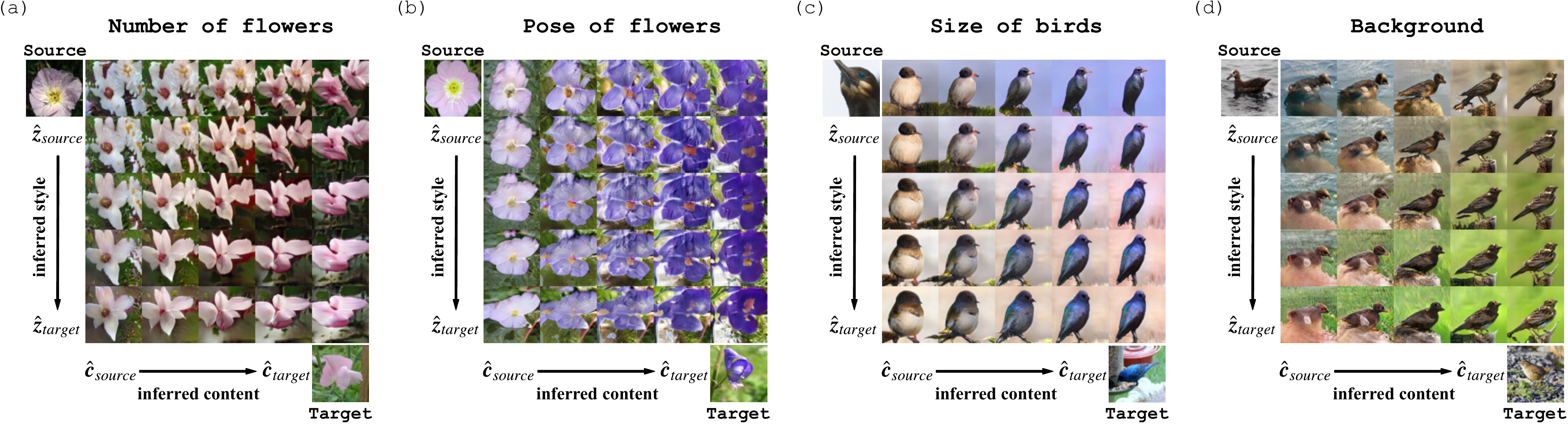}
	\caption{Examples of reconstructed images by interpolation of inferred content $\hat{\bm{c}}$ and inferred style $\hat{\bm{z}}$ from sources to targets. The learned style information includes: (a) quantity, (b) pose, (c) size and (d) background.}
	\label{fig_interpolation}
\end{figure*}

\begin{figure*}[htbp]
	\centering
	\includegraphics[width=\textwidth]{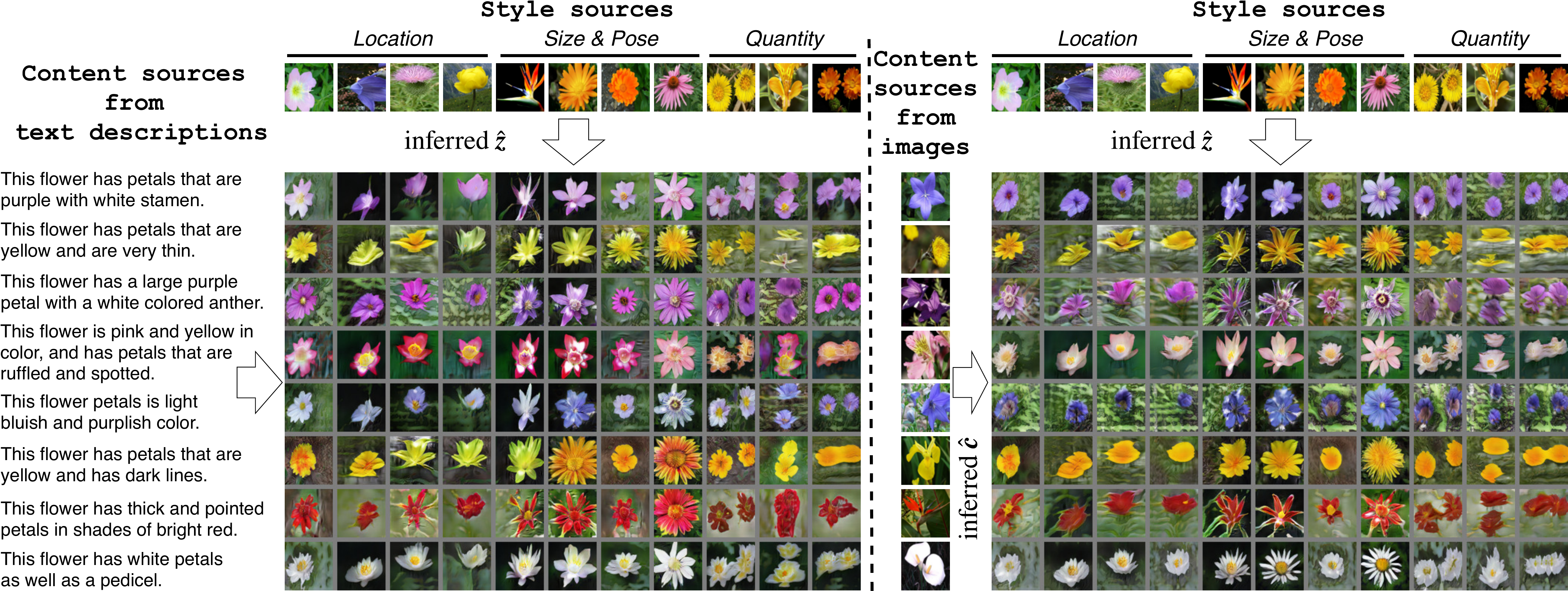}
	\caption{Disentangling content (in rows) and style (in columns) on Oxford-102 dataset by using content sources either from text descriptions (left) or images (right). More results are provided in the supplementary material (Section~\ref{sp_disentanglement}).}
	\label{fig_disentanglement_flowers}
% 	\vspace{-3mm}%Put here to reduce too much white space after your table 
\end{figure*}

\begin{figure*}[htbp]
	\centering
	\includegraphics[width=\textwidth]{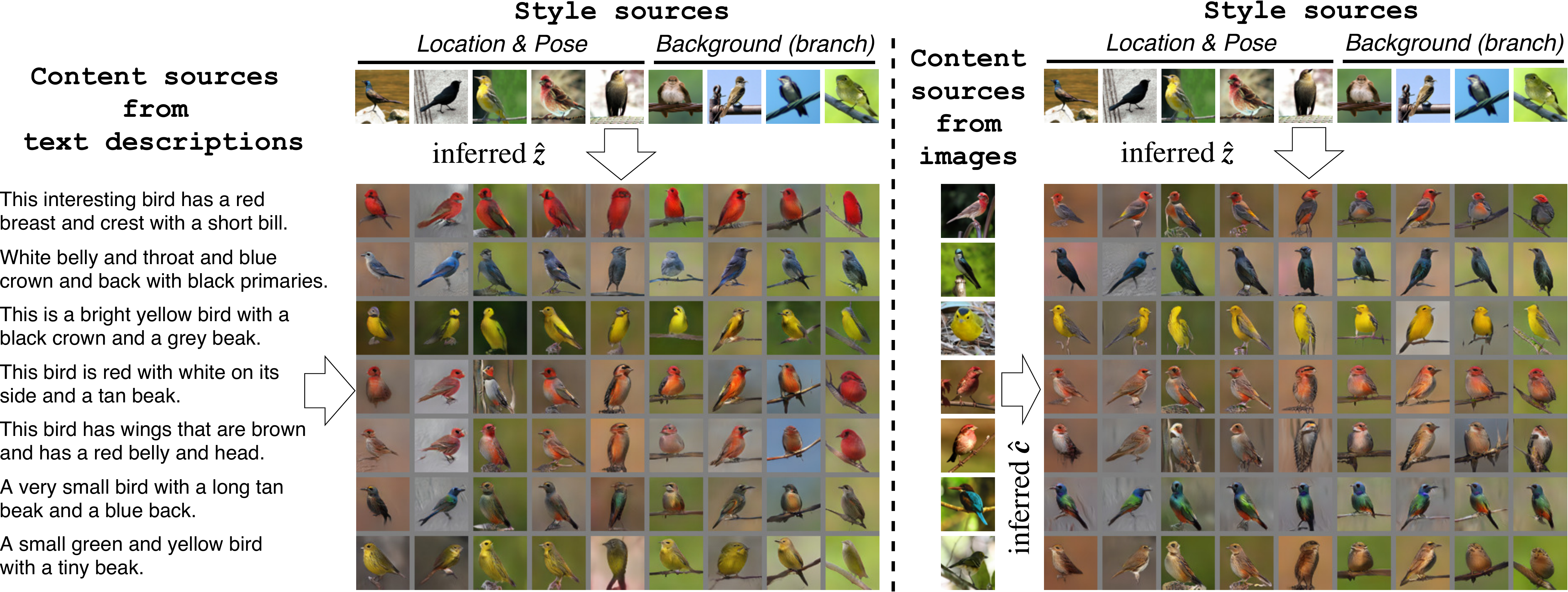}
	\caption{Disentangling content (in rows) and style (in columns) on CUB dataset by using content sources either from text descriptions (left) or images (right). More results are provided in the supplementary material (Section~\ref{sp_disentanglement}).}
	\label{fig_disentanglement_birds}
% 	\vspace{-3mm}%Put here to reduce too much white space after your table 
\end{figure*}

% \paragraph{Evaluation metrics}
% Three quantitative metrics are used to evaluate our method: Inception score \cite{salimans2016improved}, Fr\'echet inception distance (FID)~\cite{heusel2017gans} and Visual-semantic similarity~\cite{zhang2018photographic}. Inception score focuses on the diversity of generated images. 
% %with the consideration of how confident are the samples belonging to their classes. 
% We compute inception score using the fine-tuned inception models for both Oxford-102 and CUB datasets provided by \cite{zhang2017stackgan}. FID is a recently proposed metric which calculates the distance between the generated data distribution and the real data distribution, through the feature representations of the inception network. Similar to previous works~\cite{zhang2017stackgan, zhang2018photographic}, we generate $\sim$~30,000 samples when computing both inception score and FID.
% Visual-semantic similarity measures the similarity between text descriptions and generated images. We train our neural distance models for 64$\times$64 resolution images similar to ~\cite{zhang2018photographic}.

% It has been noticed in our experiments and also reported by others \cite{lucic2018gans} that, due to the variations in the training of GAN models, it is unfair to draw a conclusion based on one single experiment that achieves the best result; therefore, in our experiments, we perform three independent experiments for each method, with averages reported as final results.

\subsection{Quantitative results}
To get a global overview of how our method, the baseline method and its variants (by either fixing or removing the noise vector $\bm{z}$) behave throughout training, we evaluate each model in 20 epoch intervals. Figure~\ref{fig_is_fid} (supplementary material) shows inception score (left axis) and FID (right axis) for both Oxford-102 and CUB datasets. Consistent with the qualitative results presented in Figure~\ref{fig_problem} (supplementary material), we quantitatively show that by either fixing or removing $\bm{z}$, the baseline models retain unimpaired performance, suggesting that $\bm{z}$ has no contribution in the baseline models. % and in some cases, it may cause deterioration. This is consistent with the qualitative results we presented in Figure~\ref{fig_problem} . 
However, with our proposed dual adversarial inference, the model performance is significantly improved on FID scores for both datasets (red curves, Figure \ref{fig_is_fid}), indicating the proposed method's ability to produce better-quality images. Table \ref{tab1_performance} summarizes the comparison of the results of our method to the baseline method and also other reported results of previous state-of-the-art methods for the 64 $\times$ 64 resolution task on the three benchmark datasets: Oxford-102, CUB and COCO. %Note that although the inception score result of HDGAN \cite{zhang2018photographic} is reproducible based on one single experiment, the mean score is slight worse than the reported due to the variation.
Our method achieves the best performance based on the mean scores for both metrics on all datasets; on the FID score, it shows a 5.2\% improvement (from 40.02 to 37.94) on the Oxford-102 dataset, and a 10.6\% improvement (from 20.60 to 18.41) on the CUB dataset. In addition, we also achieve comparable results on visual-semantic similarity (Table \ref{tab_vs_similarity}, supplementary material).
%TODO: for COCO

% Moreover, given the observation of big variations not only among different experiments but also at different stages of training within the same experiment, we collect all FID scores of four different experiments evaluated at six different epochs (500, 520, 540, 560, 580 and 600) and plot the range of them for both the baseline and our method in Figure \ref{fig4_variance}. As shown in Figure \ref{fig4_variance} and also in Table \ref{tab1_performance}, the variations of the baseline method are much bigger than our method, suggesting that our method stabilizes the learning and generates more consistent and better quality images.

% \begin{figure*}[htbp]
% 	\centering
% 	\includegraphics[scale=0.28]{figures/Figure_5}
% 	\caption{Examples of reconstructed images on Oxford-102 dataset by interpolation of $\bm{z}$ (from top to bottom) and $\bm{c}$ (from left to right). The source images are shown in the top left corner and the target images are shown in the bottom right corner.}
% 	\label{fig5_recon_flower}
% \end{figure*}

\subsection{Qualitative results}
In this subsection, we present qualitative results on text-to-image generation and interpolation analysis  based on inferred content ($\hat{\bm{c}}$) and inferred style ($\hat{\bm{z}}$). 

First, we visually compare the quality and diversity of images generated from our method against the baseline. Figure \ref{fig_generation} shows one example for each dataset, illustrating that our method is able to generate better-quality images compared to the baseline method, which agrees with our quantitative results in Table \ref{tab1_performance}. We provide more examples in the supplementary material (Section~\ref{sp_generation}).  %For the Oxford-102 dataset, we generally 
%the quality of images generated by our method is significantly better than that of the baseline on both datasets. 

To make sure we are not overfitting, and to investigate whether we have learned a representative latent space, we look at interpolations of projected locations in the latent space. Interpolations also enable us to examine whether the model has indeed learned to separate style from content in an unsupervised way. %This is done by passing a pair of examples from the validation set through the encoder extract their projections in $c$ and $z$. %as ($c_1$, $z_1$) and ($c_2$, $z_2$) respectively.
%Next, we investigate the roles of the two disentangled variables $z$ and $c$ are playing in our proposed method with dual inference by the reconstruction. 
To do this, we provide the trained inference model with two images: the source image and the target image, and extract their projections $\hat{\bm{z}}$ and $\hat{\bm{c}}$ for interpolation analysis. 
As shown in Figure~\ref{fig_interpolation}, the rows correspond to reconstructed images of linear interpolations in $\hat{\bm{c}}$ from source to target image and the same for $\hat{\bm{z}}$ as displayed in columns.
The smooth transitions of both the content represented by $\hat{\bm{c}}$  from the left to right and the style represented by $\hat{\bm{z}}$  from the top to bottom indicate a good generalization of our model representing both latent spaces, and more interestingly, we find promising results showing that %in these two particular examples, 
$\hat{\bm{z}}$ is indeed controlling some meaningful style information, $\eg$, the number and pose of flowers, the size of birds and the background (Figure~\ref{fig_interpolation}, more examples in supplementary material).

\subsection{Disentanglement constraint} \label{result_dis}
%Although 
Despite promising results evidenced by many such examples as shown in Figure~\ref{fig_interpolation}, we notice that the information captured by inferred style ($\hat{\bm{z}}$) is not always consistent and faithful when we use Gaussian priors for both content and style. Inspired by the theories from independent component analysis (ICA) for separating a multivariate signal into additive subcomponents~\cite{comon1994independent}, we use a Bernoulli distribution for the content representation to satisfy the non-Gaussian constraint. This provides us with a better  %a robust 
disentanglement of content and style. Note that an alternative approach for ICA has also recently been explored in~\cite{khemakhem2019variational}. As shown in Figure~\ref{fig_disentanglement_flowers} and Figure~\ref{fig_disentanglement_birds}, our models learn to synthesize images by combining content and style information from different sources while preserving their respective properties (\eg, color for the content; and location, pose, quantity, etc. for the style), which suggests the disentanglement of content and style. Note that the content information can either directly come from a text description (left, Figure~\ref{fig_disentanglement_flowers} and Figure~\ref{fig_disentanglement_birds}) or be inferred from an image source (right, Figure~\ref{fig_disentanglement_flowers} and Figure~\ref{fig_disentanglement_birds}). More examples and discussions are provided in the supplementary material (Section~\ref{sp_disentanglement}).

Higgins~\etal~\cite{higgins2017beta} and Zhang~\etal~\cite{zhang2018separating} have proposed quantitative metrics for the disentanglement analysis which involve classification of the style attributes or comparison of the distance between generated style and true style. However, in our case, the dataset does not contain any labeled attribute that can be used to evaluate a captured style. %and styles are learned in a pure unsupervised manner,
As a result, their proposed metrics would not be suitable in our case. % (\ie, there is no labels for the style). 
One possible solution would be to artificially create a new dataset that has the same content over multiple known styles. We leave this exploration for future work.

%we conclude that in the generation phase, $\bm{z}$ learns to principally control location, pose and quantity, while the content of the object, such as color and composition, is solely controlled by $\bm{c}$. 
%However, we notice in some instances (\emph{viz.} supplementary material) that it is difficult to uncover the style information through $\bm{z}$, especially when the background is complex or the style information leaks into $\bm{c}$. This may be due to stronger emphasis on the text-matching in our current setting, and we leave further analysis for future work.

% \begin{figure*}[htbp]
% 	\centering
% 	\includegraphics[scale=0.28]{figures/Figure_6}
% 	\caption{Examples of reconstructed images on CUB dataset by interpolation of $\bm{z}$ (from top to bottom) and $\bm{c}$ (from left to right). The source images are shown in the top left corner and the target images are shown in the bottom right corner.}
% 	\label{fig6_recon_bird}
% \end{figure*}
\begin{table}[t]
% 	\begin{center}
	\centering
	    \renewcommand{\arraystretch}{1.25}
		\begin{tabular}{ccc}
			\toprule
			Method & Inception Score & FID \\
            \midrule
            ours & 3.58 $\pm$ 0.05 & 18.41 $\pm$ 1.07 \\
            ours without $V_{t2i}$ & 3.31 $\pm$ 0.04 & 20.65 $\pm$ 0.47 \\
            ours without $V_{cycle}$ & 3.53 $\pm$ 0.06 & 19.29 $\pm$ 0.90 \\
            $l_2$ loss for $V_{cycle}$ & 1.73 $\pm$ 0.15 & 149.8 $\pm$ 16.4 \\
            % ours without $V_{dual}$ \\(baseline) & 3.58 $\pm$ 0.03 & 20.60 $\pm$ 0.96 \\
			\bottomrule
		\end{tabular}
% 	\end{center}
    % \vspace{-2mm}%Put here to reduce too much white space after your table 
    \caption{Ablation study on CUB dataset. Note that the ablation on $V_{dual}$ eventually turns into the baseline.}
    \label{tab2_ablation}
    % \vspace{-4mm}%Put here to reduce too much white space after your table 
\end{table}
\subsection{Ablation study} \label{ablation_study}
In our method, we have multiple components, each of which is optimized by its corresponding objective. The previous works \cite{reed2016generative, zhang2017stackgan, zhang2018photographic} for text-to-image synthesis use the discriminator $D_{x, \varphi_t}$ to discriminate whether the image $\bm{x}$ matches its text embedding $\varphi_t$. However, with the integration of adversarial inference, where a new discriminator $D_{x,c}$ is designed to match the joint distribution of $(\bm{x}, \hat{\bm{c}})$ and $(\tilde{\bm{x}}, \bm{c})$, we now question whether the discriminator $D_{x, \varphi_t}$ is still required, given the fact that $\bm{c}$ is learned from $\varphi_t$. To answer this question, we remove the objective $V_{t2i}(D, G)$ from our method, and as seen in Table~\ref{tab2_ablation}, the performance on the CUB dataset significantly drops for both inception score and FID, indicating that $D_{x, \varphi_t}$ is not redundant in our method by providing strong supervision over the text embeddings. Similarly, we examine the role of cycle-consistency loss in our method by removing $V_{cycle}(D,G)$ from the objective. We observe a slight drop in both inception score and FID (Table \ref{tab2_ablation}), suggesting that cycle-consistency can further improve the learning of adversarial inference, which is in agreement with \cite{li2017alice}. It is also worth mentioning that our method without cycle-consistency still achieves better FID scores than the baseline method on the CUB dataset (Table~\ref{tab1_performance} and Table~\ref{tab2_ablation}), which additionally supports our proposal to integrate the inference mechanism in the current text-to-image framework.

%Instead of using adversarial loss for the cycle consistency (\ie, $\min \max(D_{x, x'}, G)$), 
We also examine the model performance by using $l_2$ loss for cycle-consistency instead of the adversarial loss. The resulting degradation in quality is unexpectedly dramatic (Table~\ref{tab2_ablation}). Figure~\ref{fig_cycleloss} (supplementary material) shows the generated images using adversarial loss compared with those using $l_2$ loss, and it is clear that the latter gives blurrier images.

%------------------------------------------------------------------------
\section{Conclusion}
In this paper, we incorporate a dual adversarial inference procedure in order to learn disentangled representations of content and style in an unsupervised way, which we show improves text-to-image synthesis. %, where two disentangled variables are simultaneously learned to represent content and style in the latent space. 
It is worth noting that the content is learned both in a supervised way through the text embedding and in an unsupervised way through the adversarial inference. The style, however, is learned solely in an unsupervised manner. Despite the challenges of the task, we show promising results on interpreting what has been learned for style. With the proposed inference mechanism, our method achieves improved quality and comparable variability in generated images evaluated on Oxford-102, CUB and COCO datasets.

% \vspace{-0.25cm}
\paragraph{Acknowledgements}
This work was supported by Mitacs project IT11934. The authors thank Nicolas Chapados for his constructive comments, and Gabriel Chartrand, Thomas Vincent, Andew Jesson, Cecile Low-Kam and Tanya Nair for their help and review.

{\small
\bibliographystyle{ieee_fullname}
\bibliography{egpaper_for_arxiv}
}

\clearpage
\onecolumn
\section{Supplementary Material}
\subsection{Problem}
\begin{figure}[hp]
	\centering
	\includegraphics[width=0.47\textwidth]{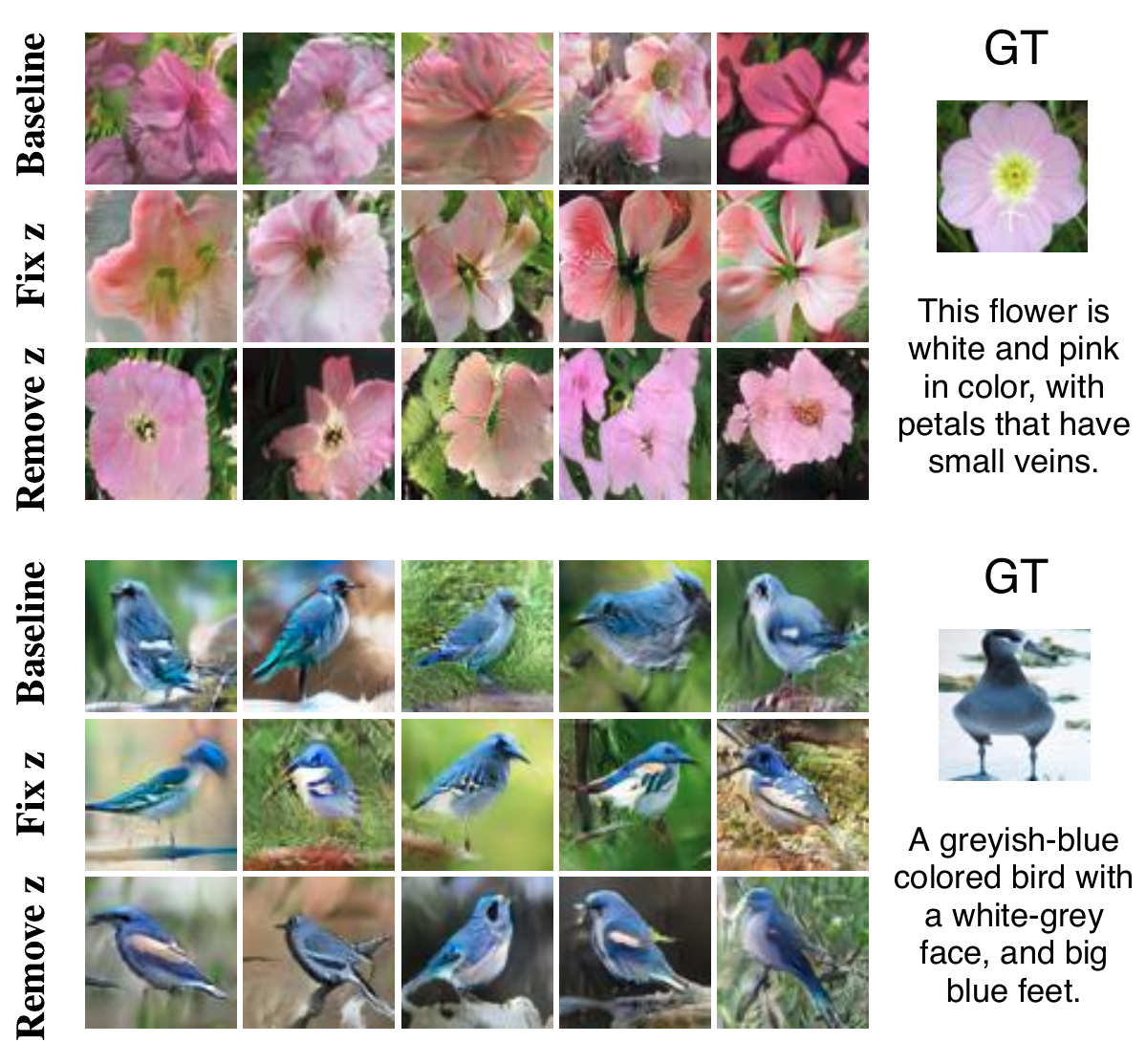}
	\caption{Generated images from previous state-of-the-art method (Baseline), fixing the noise vector $\bm{z}$ in the baseline method (Fix $\bm{z}$) and removing the noise vector $\bm{z}$ in the baseline method (Remove $\bm{z}$). The removal of the randomness from the noise source by either fixing $\bm{z}$ or removing $\bm{z}$ does not affect the variability nor the quality of generated images, indicating that the noise vector $\bm{z}$ has no contribution in the synthesis process.}
	\label{fig_problem}
\end{figure}

\begin{figure*}[htbp]
	\centering
	\includegraphics[width=\textwidth]{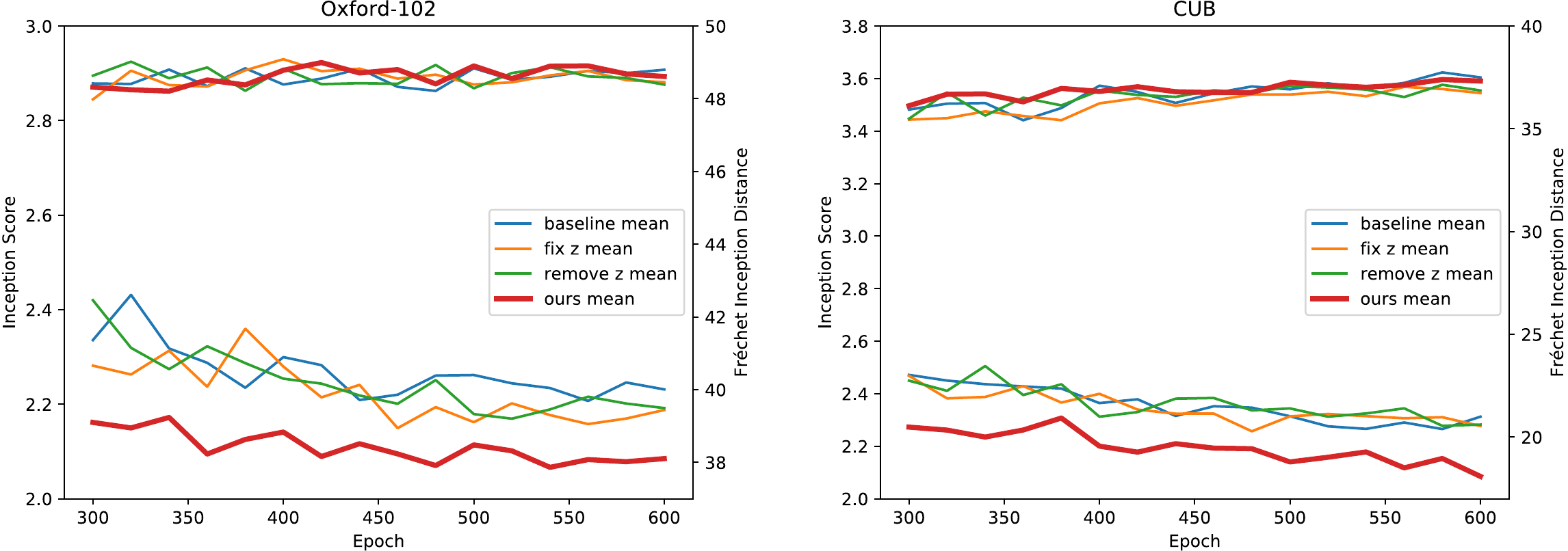}
	\caption{Inception score (left axis, top curves) and FID (right axis, bottom curves) for the baseline method, its variants (fix $\bm{z}$ and remove $\bm{z}$) and our method on Oxford-102 (left) and CUB (right) datasets. Each curve is the mean of three independent experiments. Higher inception score and lower FID mean better performance.}
	\label{fig_is_fid}
\end{figure*}

The current state-of-the-art methods for text-to-image synthesis normally have two sources of randomness: one for the text embedding variability, and the other (noise $\bm{z}$ given a normal distribution) capturing image variability. Our empirical investigation of some previously published methods reveals that those two sources can overlap: due to the randomness in the text embedding, the noise vector $\bm{z}$ then does not meaningfully contribute to the variability nor the quality of generated images, and can be discarded. This is illustrated qualitatively in Figure~\ref{fig_problem} and quantitatively in Figure~\ref{fig_is_fid}.

\subsection{Implementation details}
For the encoder $G_{z,c}$, we first extract a 1024-dimension feature vector from a given image $\bm{x}$ (note that the text embeddings are also 1024-dimension vectors), and apply the reparameterization trick to sample $\hat{\bm{z}}$ and $\hat{\bm{c}}$. To reduce the complexity of our models, we use the same discriminator for both $D_{x,z}$ and $D_{x,c}$ in our experiments, thus re-denoted as $D_{(x,z)/(x,c)}$, and the weights are shared between $D_x$ and $D_{x, \varphi_t}$. By default, $\lambda=4$. For the training, we iteratively train the discriminators $D_{x, \varphi_t}$, $D_{(x,z)/(x,c)}$, $D_{x, x'}$ and then the generators $G_x$, $G_{z,c}$ for 600 epochs (Oxford-102 and CUB) or 200 epochs (COCO), with Adam optimization \cite{kingma2015adam}. The initial learning rate is set to 0.0002 and decreased to half of the previous value for every 100 epochs.

\subsection{Datasets} 
The Oxford-102 dataset \cite{nilsback2008automated} contains 8,189 flower images from 102 different categories, and the CUB dataset \cite{welinder2010caltech} contains 11,788 bird images belonging to 200 different categories. For both datasets, each image is annotated with 10 text descriptions provided by \cite{reed2016learning}. Following the same experimental setup as used in previous works~\cite{reed2016generative, zhang2017stackgan, zhang2018photographic}, we preprocess and split both datasets into disjoint training and test sets: 82 $+$ 20 classes for the Oxford-102 dataset and 150 $+$ 50 classes for the CUB dataset. For the COCO dataset \cite{lin2014microsoft}, we use the 82,783 training images and the 40,504 validation images for our training and testing respectively, with each image given 5 text descriptions. We also use the same text embeddings pretrained by a char-CNN-RNN encoder \cite{reed2016learning}.

\subsection{Evaluation metrics}
Three quantitative metrics are used to evaluate our method: Inception score \cite{salimans2016improved}, Fr\'echet inception distance (FID)~\cite{heusel2017gans} and Visual-semantic similarity~\cite{zhang2018photographic}. Inception score focuses on the diversity of generated images. 
%with the consideration of how confident are the samples belonging to their classes. 
We compute inception score using the fine-tuned inception models for both Oxford-102 and CUB datasets provided by \cite{zhang2017stackgan}. FID is a metric which calculates the distance between the generated data distribution and the real data distribution, through the feature representations of the inception network. Similar to previous works~\cite{zhang2017stackgan, zhang2018photographic}, we generate $\sim$~30,000 samples when computing both inception score and FID.
Visual-semantic similarity measures the similarity between text descriptions and generated images. We train our neural distance models for 64$\times$64 resolution images similar to ~\cite{zhang2018photographic}.

% It has been noticed in our experiments and also reported by others \cite{lucic2018gans} that, due to the variations in the training of GAN models, it is unfair to draw a conclusion based on one single experiment that achieves the best result; therefore, in our experiments, we perform three independent experiments for each method, with averages reported as final results.

\subsection{Choices of reconstruction loss}
\begin{figure}[htbp]
	\centering
	\includegraphics[scale=0.25]{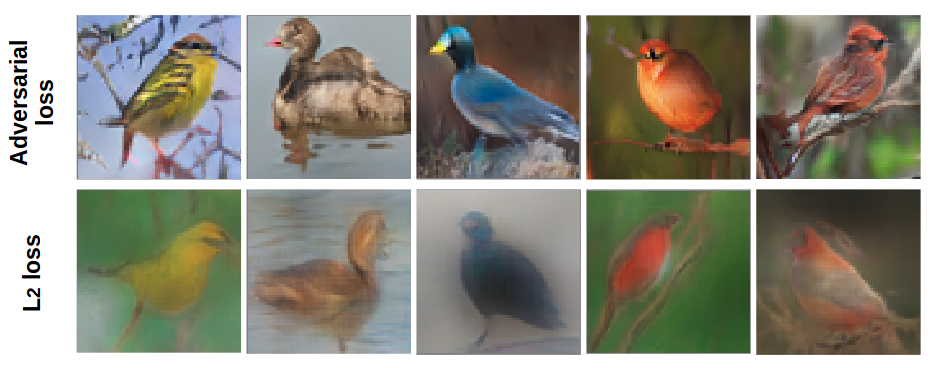}
	\caption{Examples of generated images by using either adversarial loss or $l_2$ loss for the cycle consistency.}
	\label{fig_cycleloss}
\end{figure}

In addition to the adversarial loss $V_{cycle}(D, G)$ defined in Section~\ref{sec_method_cycle}, we also examine the model performance by using either $l_1$ or $l_2$ loss for cycle-consistency. However, the degradation is unexpectedly dramatic. Figure~\ref{fig_cycleloss} shows the generated images using adversarial loss compared with those using $l_2$ loss, and it is clear that the latter gives much blurrier images. A similar effect is observed when using $l_1$ loss. The quantitative results are shown in Table~\ref{tab2_ablation}.

\subsection{Visual-semantic similarity}
Visual-semantic similarity measures the similarity between text descriptions and generated images. We train our neural distance models for 64$\times$64 resolution images similar to ~\cite{zhang2018photographic} for 500 epochs. In the testing phase, we generate $\sim$~30,000 images at 64$\times$64 resolution and compute the visual-semantic similarity scores based on the trained neural distance models. The real images are also used to compute the visual-semantic similarity scores as the ground truth. Table~\ref{tab_vs_similarity} shows the comparison of the baseline method, our method and the ground truth. Similar to inception score and FID, we provide mean scores for visual-semantic similarity metric based on three independent experiments at five different epochs (600, 580, 560, 540 and 520). As shown in the table, our method achieves comparable results for visual-semantic similarity compared to the baseline method.

% \begin{figure}
% \begin{floatrow}
% \ffigbox{%
% \includegraphics[scale=0.25]{figures/Figure_7}
% 	\caption{Examples of generated images by using either adversarial loss or $l_2$ loss for the cycle consistency.}
% 	\label{fig_cycleloss}
% }
% \capbtabbox{%
%   \begin{tabular}{cc} \hline
%   Author & Title \\ \hline
%   Knuth & The \TeX book \\
%   Lamport & \LaTeX \\ \hline
%   \end{tabular}
% }{%
%   \caption{A table}%
% }
% \end{floatrow}
% \end{figure}

% flower:
% bs: 0.24844443798065186 0.002719876437063241
% us: 0.246288775652647 0.0029878932188938953
% birds:
% bs: 0.2626403098305066 0.00435892691084229
% us: 0.2511575830479463 0.003452907752986026
% std:
% 0.13564412742853166 0.0009688158252434785
% 0.13939290270209312 0.0008274849441101997
% 0.1637516301125288 0.00085099849791885
% 0.16835698299109936 0.0011541403803732725
% Ground truth:
% flower:
% 0.4216049388051033 0.0003129536654583326
% 0.4213935285806656 0.0007250477718379031
% bird:
% 0.38183659315109253 0.0008143917540779517
% 0.3818654790520668 0.0010842952443983341
% std:
% 0.11678675934672356 0.0004486353054835967
% 0.11653710342943668 0.000569299557773606
% 0.15432864800095558 0.0007115975313972866
% 0.1541278399527073 0.0005491005902160233
\begin{table}[htbp]
	\begin{center}
% 	\scalebox{1}{%
	   \renewcommand{\arraystretch}{1.25}
		\begin{tabular}{l@{\hspace{1cm}}cc}
			\toprule
			\multirow{2}{*}{Method} & \multicolumn{2}{c}{{Dataset}} \\
			\cmidrule{2-3}
			 & Oxford-102 & CUB \\
			\midrule
			Ground Truth  & 0.422 $\pm$ 0.117 & 0.382 $\pm$ 0.154 \\
			\midrule
			HDGAN mean* & 0.248 $\pm$ 0.136 & 0.263 $\pm$ 0.164  \\
			Ours mean* & 0.246 $\pm$ 0.139 & 0.251 $\pm$ 0.168 \\
			\bottomrule
			\multicolumn{3}{l}{\footnotesize * mean calculated on three experiments at five different epochs}
		\end{tabular}
% 	}
	\end{center}
	\caption{Visual-semantic similarity.}
	\label{tab_vs_similarity}
\end{table}

% \subsection{Relation to previous work on the separation of content and style for text-to-image synthesis}
% Reed \etal \cite{reed2016generative} have also investigated the separation of content and style information. However, their learning of style is detached from the text-to-image framework, and the parameters of the image generator are fixed during the style learning phase. Their concept of content and style separation is therefore not actually leveraged during the training of the image generator. In addition, their work uses a deterministic text embedding, which cannot plausibly cover all content variations, and as a result, one can assume that information belonging to the content could severely contaminate the style. In our work, we learn style from the data itself as opposed to the generated images. This allows us to learn style while updating the generator and effectively incorporate style information from the data into the generator. 

\subsection{More examples on text-to-image generation} \label{sp_generation}
\begin{figure*}[htbp]
	\centering
	\begin{subfigure}[b]{\linewidth}
		\includegraphics[width=\linewidth]{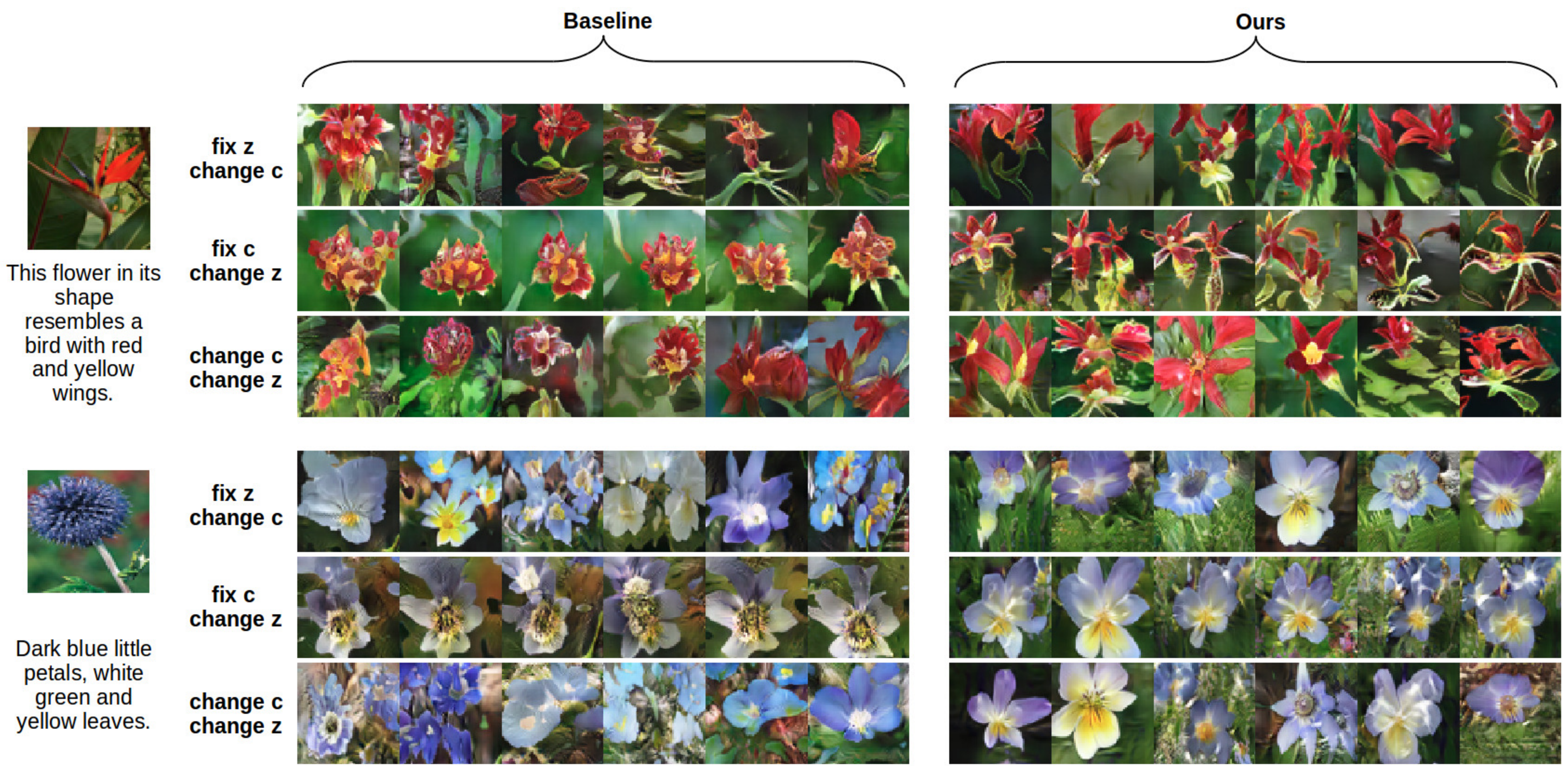}
	\end{subfigure}
	\bigskip
	\\[-4ex]
	\begin{subfigure}[b]{\linewidth}
		\includegraphics[width=\linewidth]{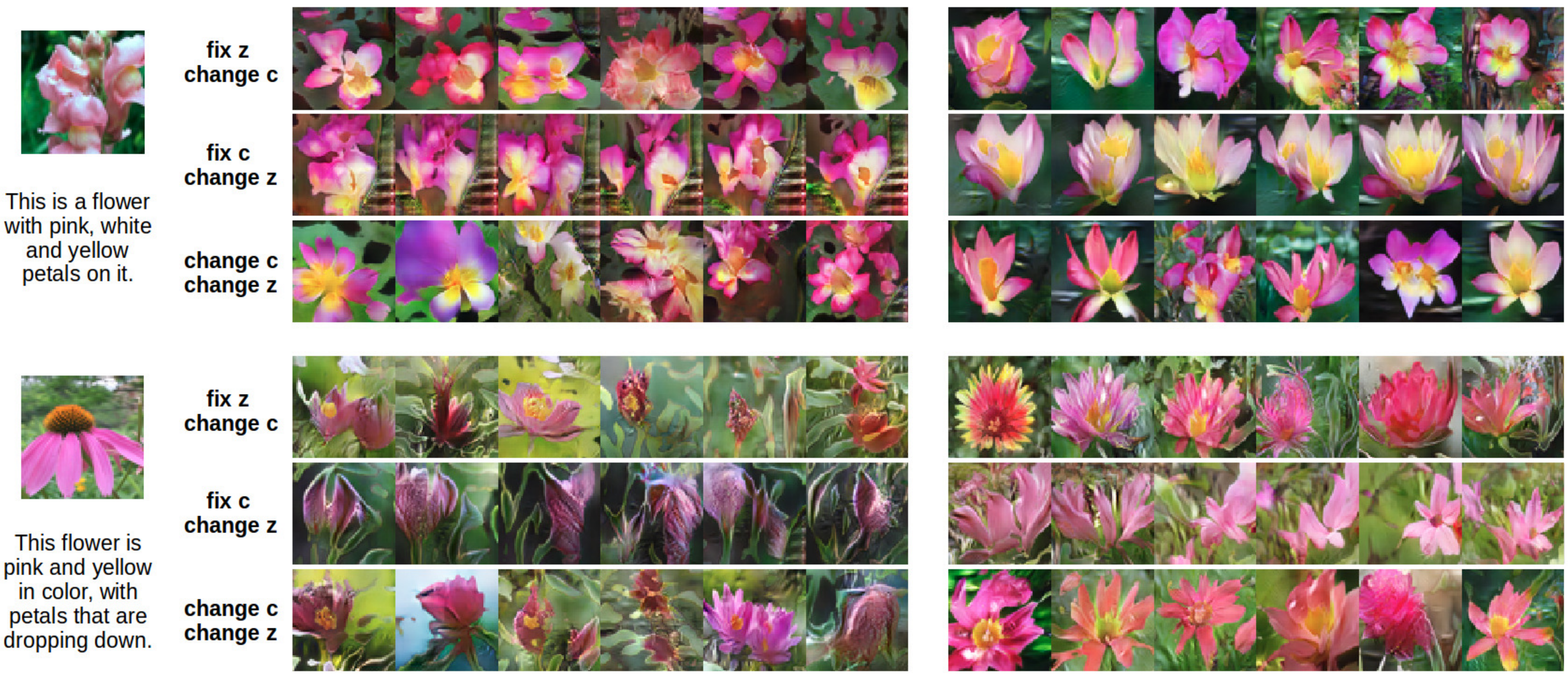}
	\end{subfigure}
	\caption{Examples of generated images on Oxford-102 dataset compared with the baseline method using three different strategies: 1) use a fixed $\bm{z}$ and sample $\bm{c}$ from $p(\bm{c})$ to show the role of $\bm{c}$ in the generated images; 2) use a fixed $\bm{c}$ and sample  $\bm{z}$ from $p(\bm{z})$ to examine how $\bm{z}$ contributes to image generation; 3) sample both $\bm{z}$ and $\bm{c}$ from $p(\bm{z})$ and $p(\bm{c})$ respectively to evaluate the overall performance. Note that $p(\bm{c})$ is conditioned on the text description. The conditioning text descriptions and their corresponding images are shown in the left column.}
	\label{sp_fig_generation}
\end{figure*}

We provide more text-to-image generation examples in Figure~\ref{sp_fig_generation}. For each method, we generate three groups of images given a certain text description: 1) use a fixed $\bm{z}$ and sample $\bm{c}$ from $p(\bm{c})$ to show the role of $\bm{c}$ in the generated images; 2) use a fixed $\bm{c}$ and sample  $\bm{z}$ from $p(\bm{z})$ to examine how $\bm{z}$ contributes to image generation; 3) sample both $\bm{z}$ and $\bm{c}$ from $p(\bm{z})$ and $p(\bm{c})$ respectively to evaluate the overall performance. Note that $p(\bm{c})$ is conditioned on the text description as defined in Section \ref{sec_methods_prelim}. As seen in Figure~\ref{sp_fig_generation}, our model generates better-quality images and gives more meaningful variability in style compared to the baseline, when we keep the content information constant and only sample from the latent variable $\bm{z}$. This is not surprising due to the fact that %we learn style representations and 
the generator has learned to match changes in $\bm{z}$ with style. 
%Note that for most of the \say{easy} tasks (images that are easy for the generator to synthesize), we do not observe significant differences between our method and the baseline by visual comparison.
Note that for most of the `easy' tasks (images that are easy for the generator to synthesize), visual comparison does not reveal significant differences between the baseline and our method.

% \clearpage
\clearpage
\subsection{More examples on disentanglement analysis} \label{sp_disentanglement}
In this subsection, we provide more results and discussions on the disentanglement analysis with the Bernoulli constraint (see Section~\ref{result_dis}). Section~\ref{sp_interpolations} gives more interpolation results based on inferred content ($\hat{\bm{c}}$) and inferred style ($\hat{\bm{z}}$); Section~\ref{sp_style_transfer} and Section~\ref{sp_style_synthetic} respectively show style transfer results based on real images and synthetic images with specifically engineered styles.

\subsubsection{Interpolations on content and style} \label{sp_interpolations}
\begin{figure*}[htbp]
	\centering
	\includegraphics[width=0.545\textwidth]{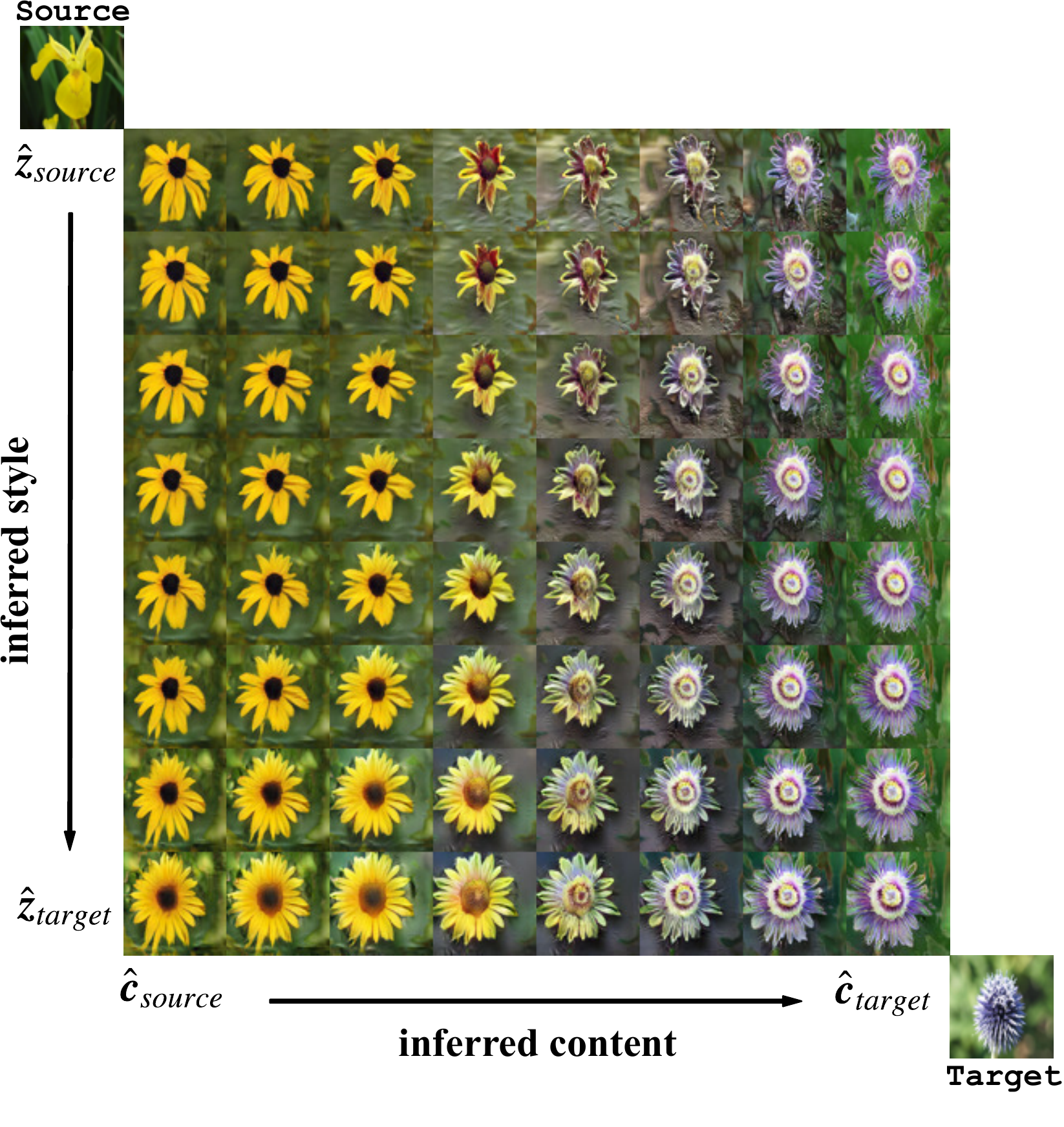}
	\caption{Example of inferred style controlling the number of petals, and the pose of flowers from facing towards upright to facing front.}
	\label{sp_fig_inter_flower1}
\end{figure*}

\begin{figure*}[htbp]
	\centering
	\includegraphics[width=0.545\textwidth]{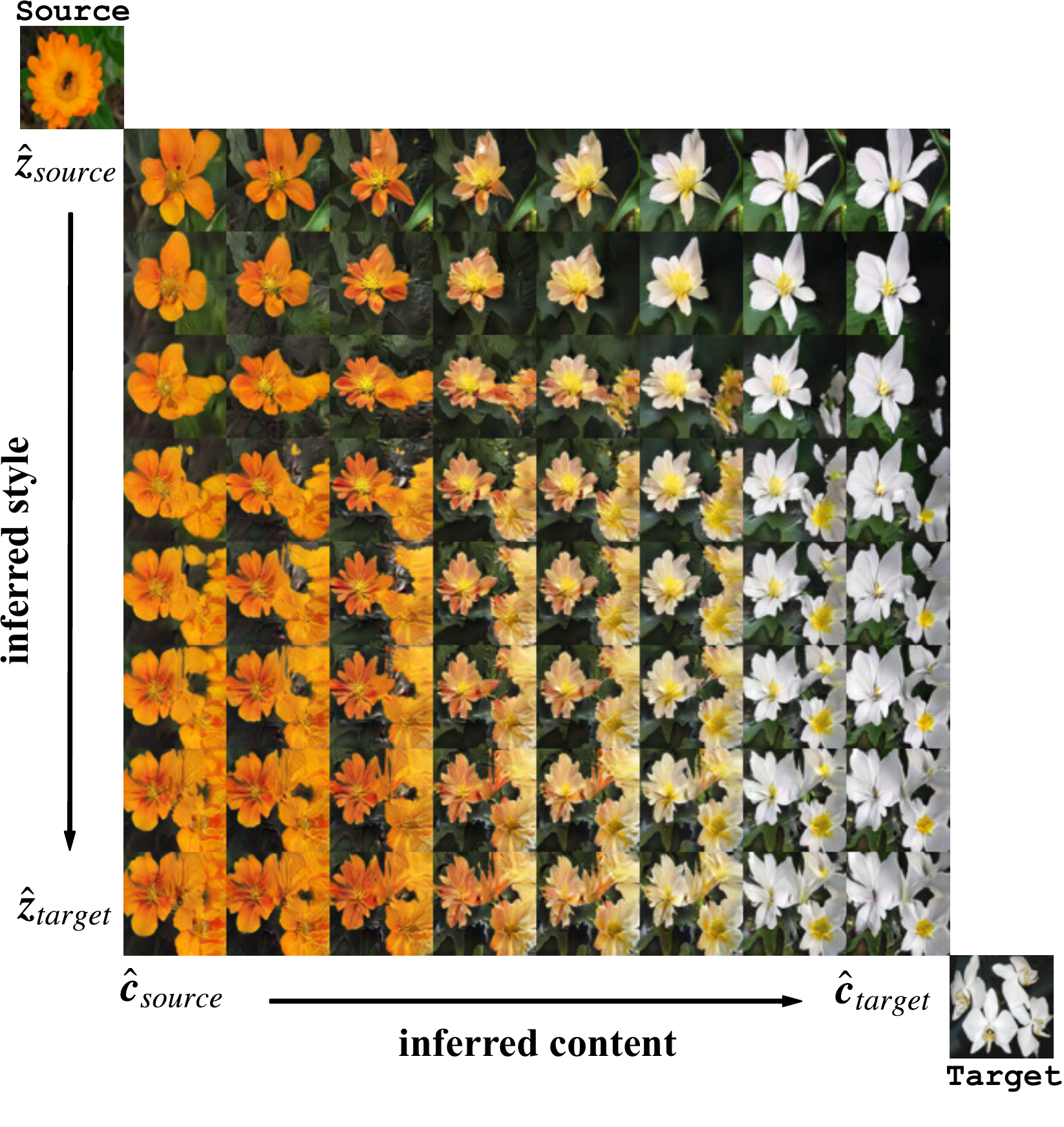}
	\caption{Example of inferred style controlling the number of flowers, from a single flower to multiple flowers.}
	\label{sp_fig_inter_flower2}
\end{figure*}

\begin{figure*}[htbp]
	\centering
	\includegraphics[width=0.545\textwidth]{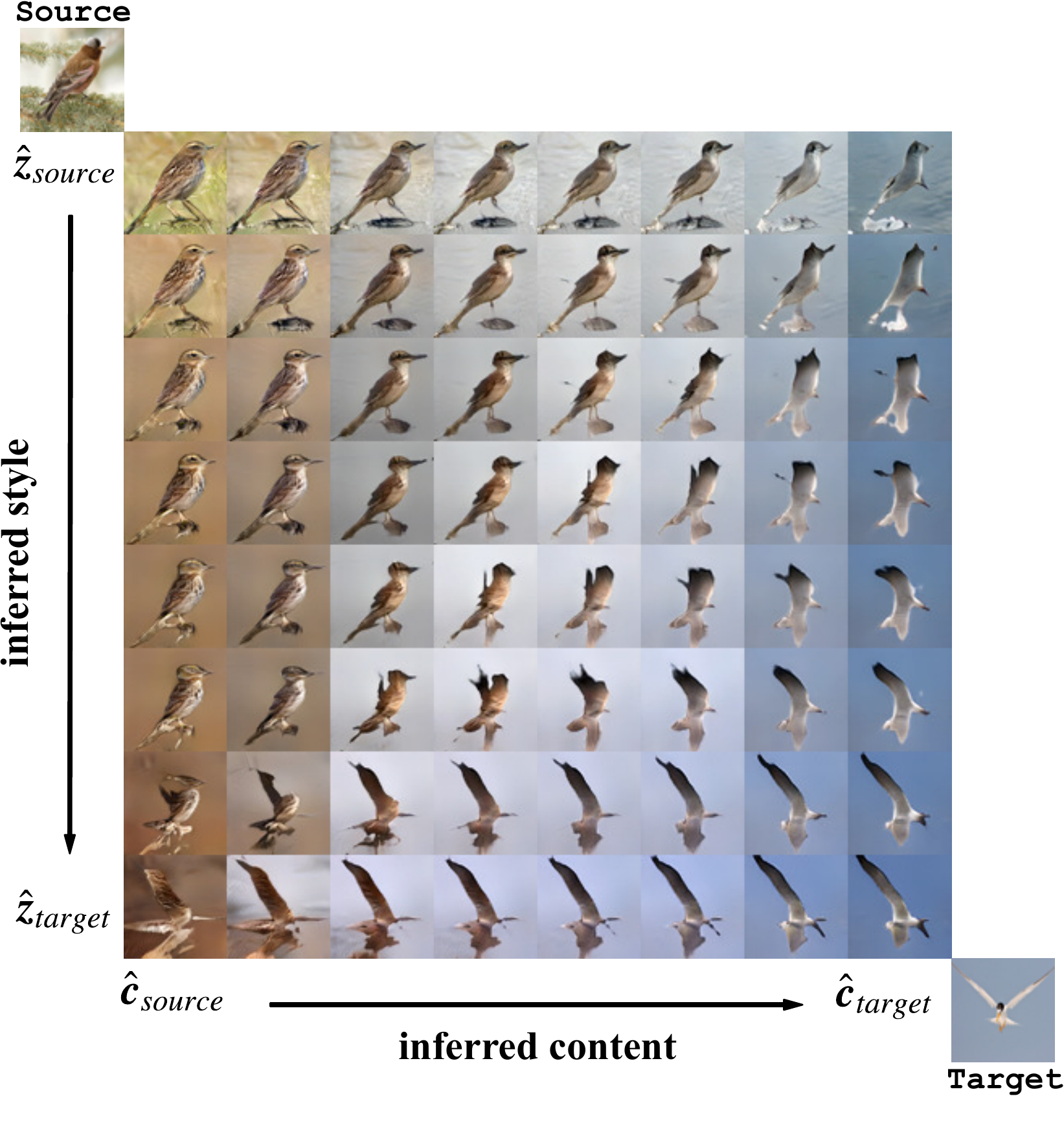}
	\caption{Example of inferred style controlling the pose of birds from sitting to flying.}
	\label{sp_fig_inter_bird1}
\end{figure*}

\begin{figure*}[htbp]
	\centering
	\includegraphics[width=0.545\textwidth]{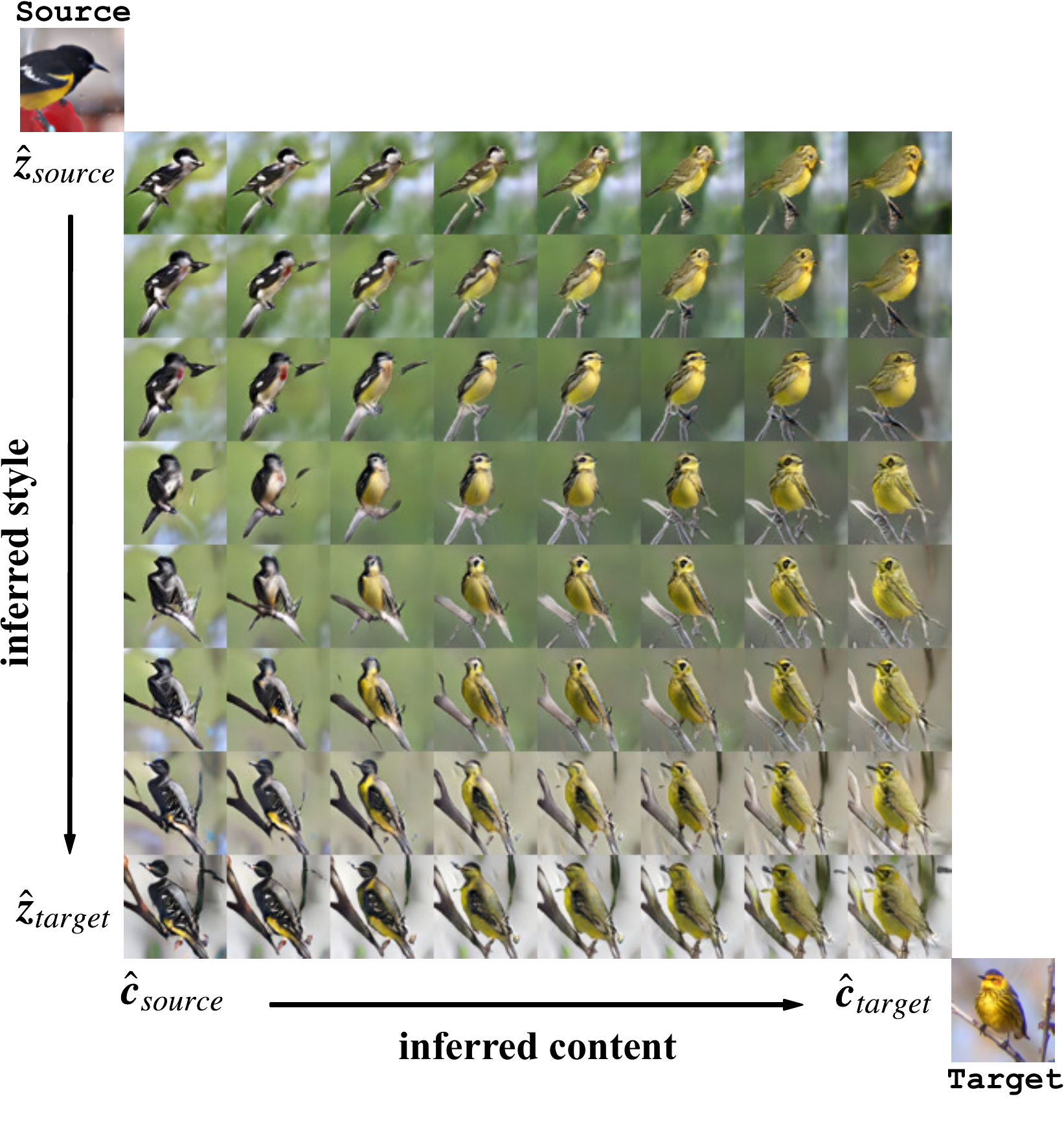}
	\caption{Example of inferred style controlling the pose of birds from facing towards right to facing towards left and the emergence of tree branches in the background.}
	\label{sp_fig_inter_bird2}
\end{figure*}

For interpolations, we provide our trained inference model with two images: the source image and the target image, to extract their projections $\hat{\bm{z}}$ and $\hat{\bm{c}}$ in the latent space. As shown in Figure~\ref{sp_fig_inter_flower1}, Figure~\ref{sp_fig_inter_flower2}, Figure~\ref{sp_fig_inter_bird1} and Figure~\ref{sp_fig_inter_bird2}, the rows correspond to reconstructed images of linear interpolations in $\hat{\bm{c}}$ from source to target image and the same for $\hat{\bm{z}}$ as displayed in columns. The source images are shown in the top left corner and the target images are shown in the bottom right corner. The figures demonstrate that our proposed dual adversarial inference is able to separate the learning of content and style, and moreover, the learned style $\hat{\bm{z}}$ indeed represents certain meaningful information. In particular, Figure~\ref{sp_fig_inter_flower1} shows an example of style controlling the number of petals, and the pose of flowers from facing towards upright to facing front; Figure~\ref{sp_fig_inter_flower2} shows a smooth transition of style from a single flower to multiple flowers; Figure~\ref{sp_fig_inter_bird1} shows the changing of the bird pose from sitting to flying; and finally Figure~\ref{sp_fig_inter_bird2} shows the switch of the bird pose from facing towards right to facing towards left and the emergence of tree branches in the background.


\clearpage

% \clearpage
\subsubsection{More style transfer results} \label{sp_style_transfer}
\begin{figure*}[htbp]
	\centering
	\includegraphics[width=\textwidth]{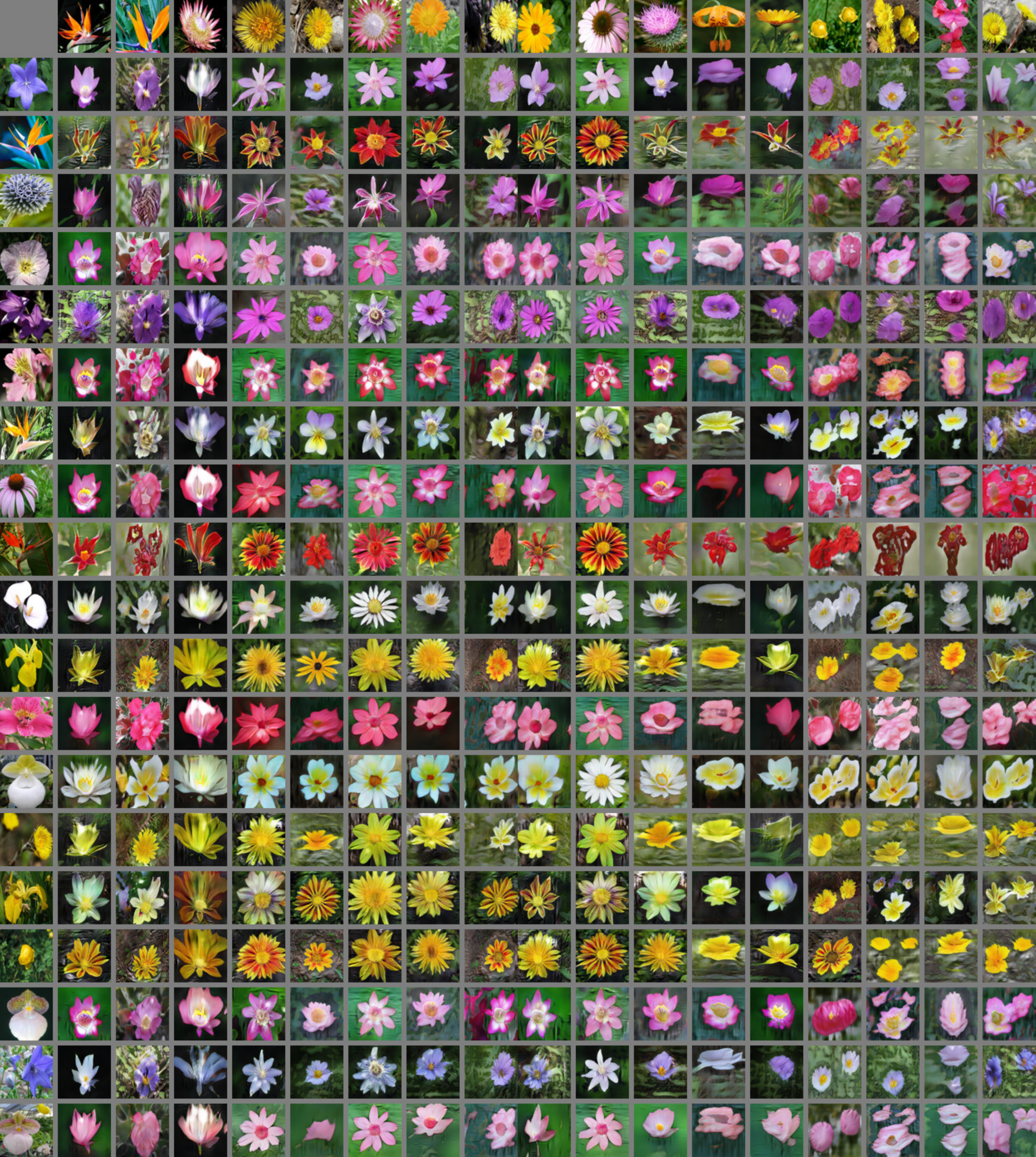}
	\caption{Disentangling content (in rows) and style (in columns) on Oxford-102 dataset by using content sources from text descriptions. The style sources are shown in the top row, and the corresponding real images for the content sources are shown in the leftmost column.}
	\label{sp_fig_dis_flower}
\end{figure*}

\begin{figure*}[htbp]
	\centering
	\includegraphics[width=\textwidth]{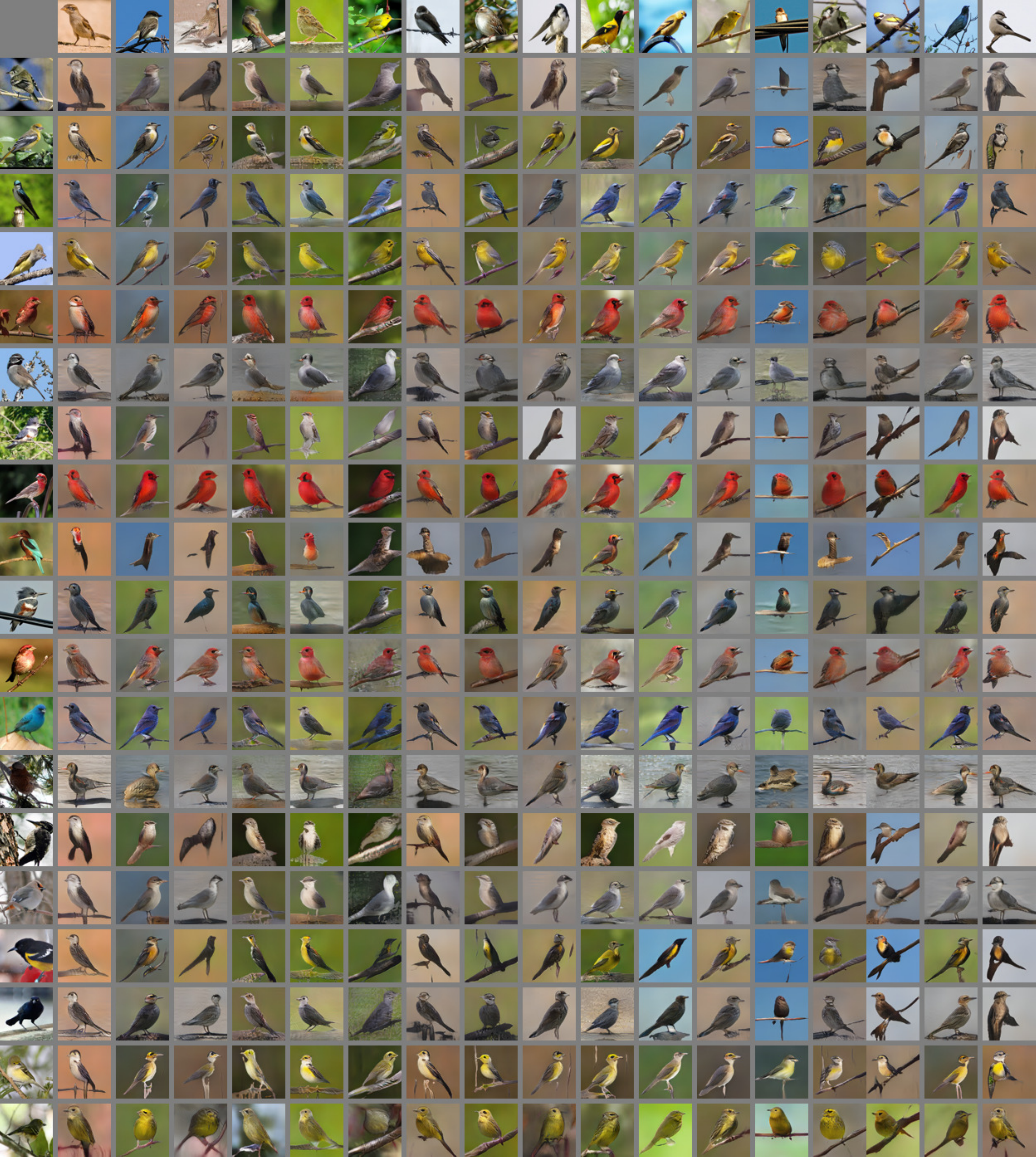}
	\caption{Disentangling content (in rows) and style (in columns) on CUB dataset by using content sources from text descriptions. The style sources are shown in the top row, and the corresponding real images for the content sources are shown in the leftmost column.}
	\label{sp_fig_dis_bird}
\end{figure*}

Here, we provide more style transfer results in Figure~\ref{sp_fig_dis_flower} (Oxford-102) and Figure~\ref{sp_fig_dis_bird} (CUB). Each column uses the same style inferred from a style source, and each row uses the same text description as the content source. The style sources that are shown in the top row, and the corresponding real images for the content sources are shown in the leftmost column.

% \subsubsection{Traversal among multiple styles}


\clearpage
\subsubsection{Style interpolations with synthetic style sources} \label{sp_style_synthetic}
\begin{figure*}[htbp]
	\centering
	\includegraphics[width=\textwidth]{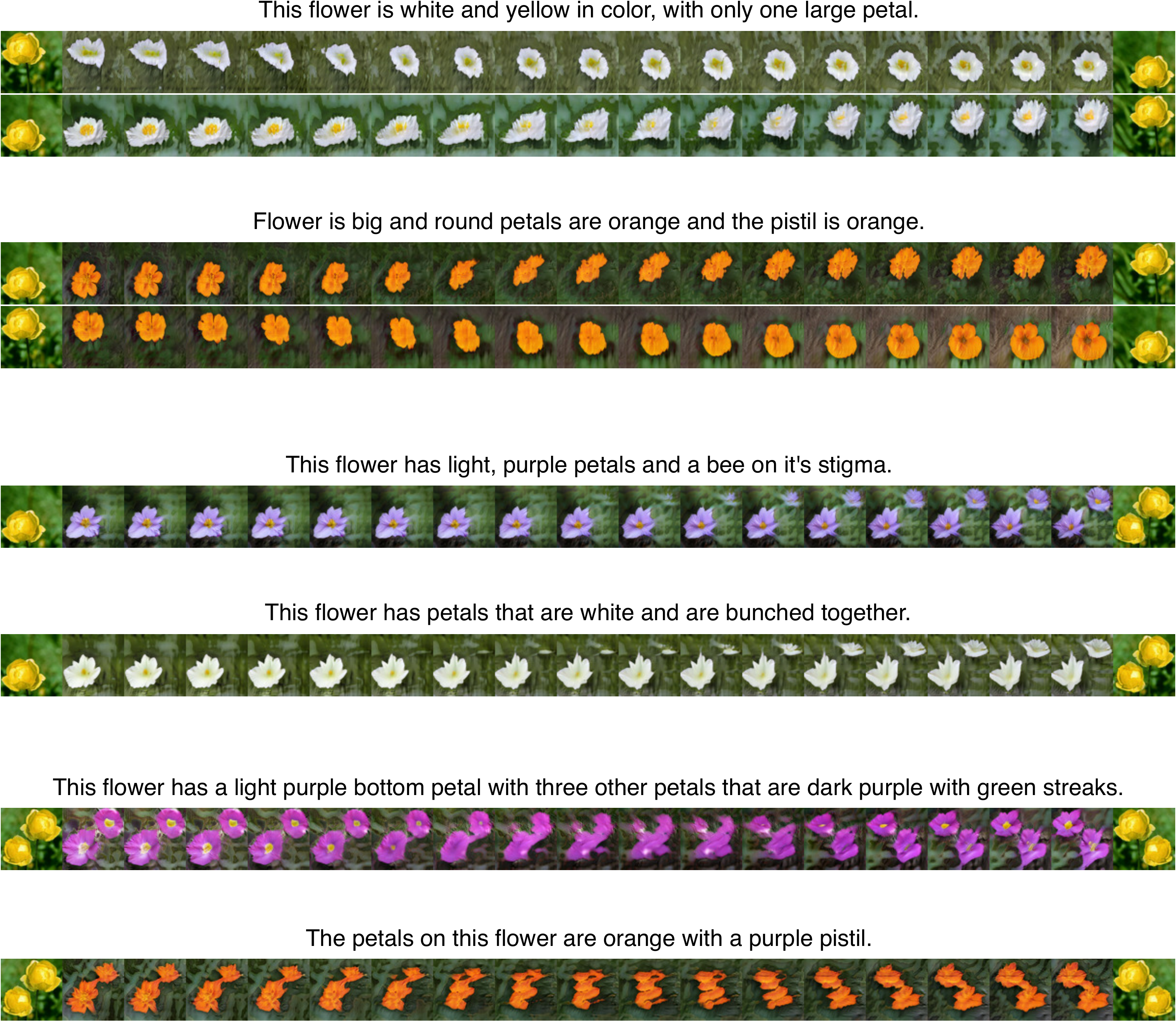}
	\caption{Style interpolations with synthetic style sources: the moving flowers and the growing quantities of flowers.}
	\label{sp_fig_synthetic}
\end{figure*}

To further validate the disentanglement of content and style, we artificially synthesize images that have certain desired known styles, \eg, a flower or multiple flowers located in different locations (top left, top right, bottom left or bottom right), and perform linear interpolation of $\hat{\bm{z}}$ from two different style sources. As shown in Figure~\ref{sp_fig_synthetic}, the flower can move smoothly from one location to another, and the number of flowers can grow smoothly from one to two, suggesting that a good representation of style information has been captured by our inference mechanism.

\end{document}